\documentclass[journal,twoside]{IEEEtran}
\usepackage{amsmath, amssymb, amsfonts}
\usepackage{url}
\usepackage{marvosym}
\usepackage{balance}
\usepackage[hidelinks,hypertexnames=false,colorlinks=true,linkcolor=blue!80!black,citecolor=red!70!black,urlcolor=black]{hyperref}
\usepackage{graphicx}
\usepackage{cite}
\usepackage{booktabs}
\usepackage{multirow}
\usepackage{threeparttable}
\usepackage{makecell}
\usepackage{mathrsfs}
\usepackage{enumitem}
\usepackage{ulem}
\usepackage{color}
\usepackage{float}
\usepackage{orcidlink}
\usepackage{amsmath}
\newcommand{\hatsup}[2]{{\smash[t]{\hat{#1}}\vphantom{#1}}^{\text{#2}}%
}
\usepackage[switch]{lineno}
\newcommand{\settablefont}{\fontsize{6}{10}\selectfont}

\newcommand{\ie}{\textit{i.e.}}

\usepackage[table,xcdraw]{xcolor}
\usepackage{scalerel}
\usepackage{tikz}
\usetikzlibrary{svg.path}

\definecolor{orcidlogocol}{HTML}{A6CE39}
\tikzset{
  orcidlogo/.pic={
    \fill[orcidlogocol] svg{M256,128c0,70.7-57.3,128-128,128C57.3,256,0,198.7,0,128C0,57.3,57.3,0,128,0C198.7,0,256,57.3,256,128z};
    \fill[white] svg{M86.3,186.2H70.9V79.1h15.4v48.4V186.2z}
                 svg{M108.9,79.1h41.6c39.6,0,57,28.3,57,53.6c0,27.5-21.5,53.6-56.8,53.6h-41.8V79.1z M124.3,172.4h24.5c34.9,0,42.9-26.5,42.9-39.7c0-21.5-13.7-39.7-43.7-39.7h-23.7V172.4z}
                 svg{M88.7,56.8c0,5.5-4.5,10.1-10.1,10.1c-5.6,0-10.1-4.6-10.1-10.1c0-5.6,4.5-10.1,10.1-10.1C84.2,46.7,88.7,51.3,88.7,56.8z};
  }
}
\newcommand\orcidicon[1]{\href{https://orcid.org/#1}{\mbox{\scalerel*{
\begin{tikzpicture}[yscale=-1,transform shape]
\pic{orcidlogo};
\end{tikzpicture}
}{|}}}}
\newcolumntype{L}[1]{>{\raggedright\arraybackslash}p{#1}}
\newcolumntype{C}[1]{>{\centering\arraybackslash}p{#1}}
\newcolumntype{R}[1]{>{\raggedleft\arraybackslash}p{#1}}

\definecolor{TYS}{rgb}{0.6, 0.8, 0.2}
\definecolor{DOcolor}{rgb}{1,0.45,0.0}
\definecolor{NAVYcolor}{rgb}{0.05,0,0.5}

\begin{document}
	
	\title{Discriminately Treating Motion Components Evolves Joint Depth and Ego-Motion Learning}
	
	\author{
		Mengtan~Zhang\,\orcidlink{0009-0003-3468-7680}, 
		Zizhan~Guo\,\orcidlink{0009-0003-7360-2192}, 
		Hongbo~Zhao\,\orcidlink{0009-0008-2198-2484}, 
		Yi~Feng\,\orcidlink{0009-0005-4885-0850}, 
		Zuyi~Xiong\,\orcidlink{0009-0001-2764-9112},
		Yue~Wang\,\orcidlink{0000-0002-0981-935X}, 
		Shaoyi~Du\,\orcidlink{0000-0002-7092-0596},
		Hanli~Wang\,\orcidlink{0000-0002-9999-4871},~\IEEEmembership{Senior Member,~IEEE}, 
		and Rui~Fan\,\orcidlink{0000-0003-2593-6596},~\IEEEmembership{Senior Member,~IEEE}
		\thanks{
			(\textit{Corresponding author: Rui Fan})}
		\thanks{Mengtan Zhang and Hongbo Zhao are with the Shanghai Research Institute for Intelligent Autonomous Systems, Tongji University, Shanghai 201210, China (e-mails: zmt200110@tongji.edu.cn, hongbozhao@tongji.edu.cn).}
		\thanks{Zizhan Guo, Yi Feng, and Zuyi Xiong are with the College of Electronic \& Information Engineering, Tongji University, Shanghai 201804, China (e-mails: 2431983@tongji.edu.cn, fengyi@ieee.org, 2153478@tongji.edu.cn).}
		\thanks{Yue Wang is with the Department of Control Science and Engineering, Zhejiang University, Hangzhou, Zhejiang 310027, China (e-mail: wangyue@iipc.zju.edu.cn).}
		\thanks{Shaoyi Du is with the National Key Laboratory of Human-Machine Hybrid Augmented Intelligence, the National Engineering Research Center for Visual Information and Applications, and the Institute of Artificial Intelligence and Robotics, Xi'an Jiaotong University, Xi’an, Shaanxi 710049, China (e-mail: dushaoyi@xjtu.edu.cn).}
		\thanks{Hanli Wang is with the College of Electronic \& Information Engineering, the School of Computer Science and Technology, and the Key Laboratory of Embedded System and Service Computing (Ministry of Education), Tongji University, Shanghai 201804, China (e-mail: hanliwang@tongji.edu.cn).}
		\thanks{Rui Fan is with the College of Electronic \& Information Engineering, Shanghai Institute of Intelligent Science and Technology, Shanghai Research Institute for Intelligent Autonomous Systems, the State Key Laboratory of Autonomous Intelligent Unmanned Systems, the Frontiers Science Center for Intelligent Autonomous Systems (Ministry of Education), and Shanghai Key Laboratory of Intelligent Autonomous Systems, Tongji University, Shanghai 201804, China, as well as with the National Key Laboratory of Human-Machine Hybrid Augmented Intelligence, Xi'an Jiaotong University, Xi'An, Shaanxi 710049, China (e-mail: rui.fan@ieee.org).}
	}

	\maketitle
	
		\begin{abstract}
			Unsupervised learning of depth and ego-motion, two fundamental tasks in 3D perception, has made significant strides in recent years. Nevertheless, most existing methods often treat ego-motion estimation as an auxiliary task, either indiscriminately mixing all motion types or excluding depth-independent rotational motions when generating supervisory signals. Such designs hinder the incorporation of sufficiently strong geometric constraints into the joint learning framework, thereby limiting the reliability and robustness of these models under diverse environmental conditions.
			This study introduces a discriminative treatment of motion components by leveraging the distinct geometric regularities inherent in their respective rigid flows, which simultaneously benefits both depth and ego-motion estimation. Given consecutive video frames, the network outputs are first employed to align the optical axes and imaging planes of the source and target cameras. Optical flows between the original frames are then transformed through these alignment processes, and their deviations are quantified to impose geometric constraints. These constraints are applied individually to each estimated ego-motion component, enabling more targeted and reliable refinement. These alignments further reformulate the joint learning process into coaxial and coplanar forms, where depth and each translation component can be mutually derived through closed-form geometric relationships, introducing complementary constraints that significantly improve the robustness of depth learning.
			DiMoDE, a general depth and ego-motion joint learning framework that incorporates all these innovative designs, achieves state-of-the-art performance in both tasks, as demonstrated by extensive experiments on multiple public datasets and a newly collected, diverse real-world dataset, particularly under challenging conditions where existing approaches often fail. Our source code will be publicly available at \href{https://mias.group/DiMoDE}{mias.group/DiMoDE} upon publication.
		\end{abstract}
		\begin{IEEEkeywords}
			depth estimation, visual odometry, motion component, geometric constraint.
		\end{IEEEkeywords}

	\section{Introduction}
	\label{sect.introduction}
	
	\IEEEPARstart{A}{ccurate} estimation of ego-motion and depth is crucial for 3D perception \cite{geiger2013vision}. Recently, there has been a significant increase in methods that jointly learn ego-motion and depth from monocular video sequences in an unsupervised, end-to-end manner \cite{godard2019digging, sun2023sc, zhang2023lite}.
	Compared to supervised depth estimation approaches \cite{eigen2014depth,shao2024nddepth, yang2024depth}, this paradigm eliminates the need for extensive, manually labeled annotations \cite{zhou2017unsupervised, godard2019digging}, which are not always available in real-world scenes. Moreover, compared to traditional visual odometry methods that rely on 2D visual correspondences to explicitly solve the relative pose problem, such a joint learning paradigm can maintain robustness in scenarios where 2D correspondences are fairly sparse or unreliable \cite{yang2018challenges, zou2020learning}.
	
	This unsupervised joint learning paradigm typically consists of a depth estimation network (hereafter known as \textit{DepthNet}) and a pose estimation network (hereafter known as \textit{PoseNet}) \cite{zhou2017unsupervised,zhang2023lite}. DepthNet estimates a dense depth map. PoseNet regresses the following ego-motion transformation:
	\begin{equation}
		\boldsymbol{T}=
		\begin{pmatrix}
			\boldsymbol{R} & \boldsymbol{t}
			\\
			\boldsymbol{0}^\top & 1
		\end{pmatrix} 
		\in \mathrm{SE(3)},
	\end{equation}
	where $\boldsymbol{R} \in \mathrm{SO(3)}$ denotes the rotation matrix, $\boldsymbol{t}=(t_x, t_y, t_z)^\top \in \mathbb{R}^3$ denotes the translation vector, and $\boldsymbol{0}$ denotes a column vector of zeros. The estimated depth map and ego-motion are then utilized to calculate the rigid flow, a key component for warping the source image into the target view. By minimizing the photometric difference between the target and the warped images, these two networks can be trained simultaneously. 
	
	\begin{figure*}[t!]
		\centering
		\includegraphics[width=0.95\textwidth]{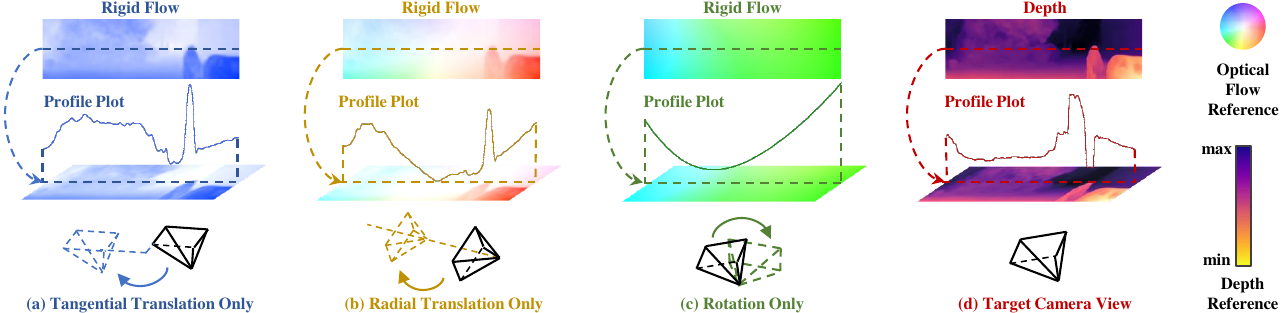}
		\caption{Tangential and radial translations result in geometrically regular but distinct depth-dependent flows. Specifically, the rigid flow induced by tangential translation varies inversely with depth, while the one resulting from radial translation is not only depth-dependent but also subject to perspective scaling. In contrast, rotation results in irregular, depth-independent flows.
		}
		\label{fig.motivation}
	\end{figure*}
	
	Fig. \ref{fig.motivation} presents the rigid flows resulting from various motion types, a topic that has received limited discussion in previous studies. As shown in Figs. \ref{fig.motivation}(a) and \ref{fig.motivation}(b), ego-motion that involves only translations yields fairly regular rigid flows. Specifically, when the camera moves perpendicularly to the optical axis, leading to a tangential translation vector $\boldsymbol{t}^{\text{Tan}}=(t_x, t_y, 0)^\top$, the rigid flows of all pixels are parallel. When the camera moves along the optical axis, leading to a radial translation vector $\boldsymbol{t}^{\text{Rad}}=(0, 0, t_z)^\top$, the rigid flows of all pixels move toward or away from the principal point. These regular rigid flows provide vital visual clues that can potentially improve the accuracy of estimated motion components. On the other hand, as illustrated in Fig. \ref{fig.motivation}(c), ego-motion that involves only rotations results in fairly irregular rigid flows. Therefore, it is inappropriate to train PoseNet to indiscriminately regress rotation and translation components by mixing these two types of rigid flows for supervisory signal generation. However, there are currently few previous studies that focus specifically on this aspect. 
	
	In addition, we revisit Fig. \ref{fig.motivation} to explore potential improvements in DepthNet training. As demonstrated in the study \cite{bian2021auto}, two images related by a pure-rotation transformation can be warped from one to the other using a homography matrix $\boldsymbol{H}=\boldsymbol{K}\boldsymbol{R}\boldsymbol{K}^{-1}$, where $\boldsymbol{K}$ denotes the camera's intrinsic matrix. Therefore, as shown in Fig. \ref{fig.motivation}(c), the rigid flow produced solely by rotation is independent of depth and does not provide effective gradients necessary for the supervision of DepthNet training. In contrast, as illustrated in Fig. \ref{fig.motivation}(a), pure tangential translation results in rigid flows characterized by superposed horizontal and vertical positional shifts. These shifts, commonly referred to as disparities in stereo vision, are inversely proportional to depth (see Fig. \ref{fig.motivation}(d)) and can directly contribute to the training of DepthNet. Furthermore, as illustrated in Fig. \ref{fig.motivation}(b), pure radial translation leads to perspective scaling in sequential images, which is also depth-dependent. Specifically, the scale change of an object in the image is inversely proportional to its depth. Thus, these rigid flows also provide valuable visual cues that can be effectively incorporated into the DepthNet training process. However, perspective scaling causes pixels near the image boundary to undergo larger displacements than those near the principal point. Consequently, mixing these two types of visual cues can lead to inconsistent gradients across the image, potentially impeding convergence and degrading depth estimation performance. Despite this limitation, the significant differences between tangential and radial translations have long been overlooked, with existing studies generally treating them indiscriminately.
	
	These observations motivate us to discriminately treat the motion components in $\boldsymbol{T}$ by isolating the rotation and separately processing the tangential and radial translations. Specifically, the estimated rotation $\boldsymbol{R}$ is expected to align the camera orientation of the source view with that of the target view, thereby eliminating irregular rotational flows. The estimated tangential translation $\boldsymbol{t}^{\text{Tan}}$ and radial translation $\boldsymbol{t}^{\text{Rad}}$ are then expected to align the camera's optical axes and imaging planes in the source and target views, respectively, thereby generating two types of regular flows containing vital visual clues. 
	
	Building upon the aforementioned analyses, this study aims to reformulate the existing joint learning framework for depth and ego-motion estimation by introducing explicit geometric constraints from motion components. These constraints are imposed through the alignment of the optical axes and imaging planes between the source and target cameras. In practice, we utilize dense correspondences that adhere to the static scene hypothesis to objectively quantify the extent of both types of alignment. These correspondences, when transformed by performing both alignments based on the estimated depth and ego-motion, are expected to exhibit the aforementioned geometric regularities. Deviations from the expected geometric regularities are employed as supervisory signals to independently guide the learning of tangential and radial translations, while simultaneously constraining the learning of rotation. As the transformed correspondences progressively exhibit expected geometric regularities, depth-dependent visual cues arising from positional shifts and perspective scaling are effectively disentangled. By exploiting their distinct depth-dependent variations, these visual cues are processed separately to recover depth, thereby providing per-pixel constraints for depth estimation. The depth estimated under these constraints and the transformed correspondences are subsequently used to compute tangential and radial translations based on simplified geometric relationships. A geometric constraint is also incorporated to enforce consistency between these computed translations and those predicted by the PoseNet, thereby establishing a constraining cycle that enhances the robustness of depth learning. Incorporating the above-mentioned innovative components into a general depth and ego-motion joint learning framework yields \textbf{DiMoDE}, which discriminately treats motion components. Extensive experiments on multiple public datasets and our collected real-world video sequences demonstrate the state-of-the-art (SoTA) performance of our proposed DiMoDE in both monocular depth estimation and visual odometry, as well as the efficacy of discriminately treating motion components.
	
	In summary, our main contributions are as follows:
	\begin{itemize}
		\item 
		We conduct both phenomenological and theoretical analyses on the distinctions among rigid flows arising from different motion types and the long-overlooked limitations caused by indiscriminately mixing these flows when generating supervisory signals.
		
		\item 
		We introduce optical axis and imaging plane alignment processes that reformulate joint learning from monocular video into coaxial and coplanar forms, thereby mitigating the effects of mixed motion types and enhancing the robustness and stability of depth learning.
		
		\item 
		We propose a discriminative treatment of motion components that enables these alignments to be achieved using network outputs, not only eliminating the need for auxiliary pose estimation algorithms but also refining each component of the PoseNet outputs.
		
		\item 
		We develop DiMoDE, a general framework compatible with various models, achieving notable improvements in network convergence and training robustness for both depth estimation and visual odometry across diverse challenging conditions.
		
	\end{itemize}
	
	The remainder of this article is organized as follows:
	Sect. \ref{Sect.related_work} reviews existing monocular visual odometry and depth estimation methods. 
	Sect. \ref{sect.method0} introduces preliminaries of monocular depth and ego-motion joint learning.
	Sect.~\ref{sect.method1} details the proposed discriminative treatment of motion components, and Sect.~\ref{sect.method2} presents the resulting DiMoDE framework, respectively.
	Sect. \ref{Sect.experiments} presents the experimental results across several public datasets and our collected real-world dataset. 
	In Sect. \ref{Sect.discussion}, we discuss two limitations of the DiMoDE framework. 
	Finally, Sect. \ref{Sect.conclusion} concludes this article and discusses possible directions for future work.

	\section{Related Work}
	\label{Sect.related_work}

	\subsection{Monocular Visual Odometry}
	\label{sec.geometry_vo_related}
	
	Monocular visual odometry is a fundamental and extensively studied problem in the fields of robotics and computer vision \cite{zhan2020visual}. Existing approaches can be broadly categorized into three main classes: geometry-based, learning-based, and hybrid.
	
	Earlier geometry-based methods, such as VISO2 \cite{geiger2011stereoscan} and ORB-SLAM series \cite{mur-artal2015orbslam, mur2017orb, campos2021orbslam3}, primarily rely on hand-crafted keypoints and descriptors for correspondence matching, and often perform poorly in texture-less environments \cite{zhan2020visual, yang2018challenges}. To address this limitation, methods like DSO \cite{engel2017direct} formulate direct image or feature warping as an energy minimization problem, thus avoiding the need for accurate correspondence matching. Despite their robustness in texture-less environments, these methods exhibit a lower accuracy ceiling compared to approaches based on correspondence matching \cite{sun2022improving}.
	
	In recent years, extensive learning-based methods \cite{wang2017deepvo, shen2019beyond} have been proposed to tackle the above-mentioned challenges. These methods typically employ a PoseNet to regress camera poses from consecutive video frames in an end-to-end manner \cite{wang2018endend, li2018undeepvo, zhou2017unsupervised, bian2021unsupervised}. By eliminating the need for explicit correspondence matching used to solve for relative pose, PoseNet remains robust in texture-less scenes and typically supports efficient inference. Among them, the study \cite{zhou2017unsupervised} proposes a joint depth and ego-motion learning paradigm that trains the PoseNet with only unlabeled monocular video, thereby alleviating the need for expensive manual annotation. Subsequent studies have focused on addressing issues such as scale inconsistency and limited generalizability \cite{bian2021unsupervised, zhang2022information}, while also incorporating temporal and spatial cues into PoseNet to improve their performance \cite{zou2020learning, jiang2022mlfvo, feng2024scipad}. Despite these advancements, the fundamental challenges of generating more reliable supervisory signals in an unsupervised manner to ensure accurate and robust ego-motion learning remain unsolved \cite{zhan2020visual, sun2022improving}. 
	
	More recently, hybrid approaches \cite{yang2020d3vo, tiwari2020pseudo, zhan2020visual, sun2022improving} have introduced a new paradigm that integrates neural network predictions, such as depth and correspondence matching, with back-end optimization for relative pose estimation. Despite the incorporation of geometric constraints during inference, enabling these hybrid methods to achieve superior performance over fully learning-based approaches, the resulting high storage demands and computational complexity significantly hinder their applicability in real-world, resource-constrained environments.
	
	This study incorporates explicit geometric constraints from each ego-motion component by enforcing the alignment of optical axes and imaging planes within the joint learning framework. The proposed DiMoDE framework enables a simple ResNet-based PoseNet \cite{he2016deep} to achieve performance comparable to SoTA geometric-based and hybrid approaches, with significantly lower storage and computational overhead.

	\subsection{Monocular Depth Estimation}
	\label{sec.unsupervised_related}
	
	Monocular depth estimation, which infers depth information from a single RGB image, has found widespread applications in areas such as autonomous driving and embodied artificial intelligence. Early approaches typically adopted a supervised learning paradigm that requires extensive depth ground truth for model training \cite{eigen2014depth}. Building on this foundation, subsequent studies \cite{cao2017estimating, fu2018deep, bhat2021adabins, shao2023iebins} improved depth estimation performance by reformulating the depth regression problem as either a per-pixel classification task that predicts discretized depth intervals or a classification-regression task that predicts depth as a weighted combination of bin centers. More recently, the studies \cite{yang2024depth, yang2025depth, hu2024metric3d} have leveraged vision foundation models to achieve highly accurate depth estimation.
	
	To alleviate the need for extensive ground-truth data in monocular depth estimation, growing interest has been directed toward unsupervised learning of depth from monocular video, a paradigm first introduced in the study \cite{zhou2017unsupervised}. To improve depth estimation performance within this paradigm, the study \cite{godard2019digging} introduced a per-pixel minimum reprojection loss to enhance the photometric supervisory signals, while studies \cite{he2022ra, zhang2023lite, zhou2024recurrent} explored more sophisticated network architectures. Subsequent studies \cite{zou2018df, sun2023sc, sun2023dynamo, shao2024monodiffusion, zhang2024dcpi} have sought to preserve the effectiveness of this training paradigm in scenarios with occlusions and dynamic objects by introducing additional visual cues. However, these methods rely solely on per-pixel supervisory consistency, which limits their robustness under adverse conditions, such as nighttime or poor weather, where such signals often become unreliable.
	
	Recently, several studies \cite{yan2025synthetic, sun2025depth} have focused specifically on enhancing the robustness of depth estimation under adverse conditions. These methods typically adopt domain adaptation strategies, wherein networks are initially trained on clear daytime scenes and subsequently adapted to more challenging environments such as nighttime or fog. Consequently, their reliance on fixed data splits hinders the full utilization of large-scale data collected under diverse and dynamically changing conditions, limiting the continuous evolution of DepthNet.
	
	In contrast, this study focuses on the discriminative treatment of motion components during the original supervisory signal generation, enabling a robust constraint cycle for depth estimation. To the best of our knowledge, DiMoDE is the first approach to effectively address the challenges posed by adverse environments within a unified and general joint learning framework.

	\section{Preliminaries} 
	\label{sect.method0}
	Given a monocular camera with the intrinsic matrix
	\begin{equation}
		\label{eq.intrinsic_matrix}
		\boldsymbol{K} = \begin{pmatrix}
			f_u&0&u_0 \\0& f_v& v_0\\0&0&1
		\end{pmatrix},
	\end{equation}
	where $f_u$ and $f_v$ denote the focal lengths in the horizontal and vertical directions, respectively, and $\boldsymbol{p}_0 = (u_0, v_0)^\top$ represents the principal point, a pair of RGB images, $\boldsymbol{I}_{t}$ and $\boldsymbol{I}_{s}$, captured from the target and source views, respectively, are fed into the DepthNet to generate their corresponding depth maps, $\boldsymbol{D}_{t}$ and $\boldsymbol{D}_{s}$. Let $\boldsymbol{p}_t=(u_t, v_t)^\top$ and $\boldsymbol{p}_s=(u_s, v_s)^\top$ be a pair of corresponding pixels in the target and source images, respectively. The depth values at these pixels are denoted as $\hat{z}_{t}=\boldsymbol{D}_{t}(\boldsymbol{p}_{t})$ and $\hat{z}_{s}=\boldsymbol{D}_{s}(\boldsymbol{p}_{s})$. Their respective homogeneous coordinates are denoted as $\tilde{\boldsymbol{p}}_{t}$ and $\tilde{\boldsymbol{p}}_{s}$. Meanwhile, $\boldsymbol{I}_{t}$ and $\boldsymbol{I}_{s}$ are concatenated and fed into the PoseNet to estimate the ego-motion $\hat{\boldsymbol{T}}$ from the source view to the target view. The estimated depth and ego-motion are then used to generate a rigid flow map $\boldsymbol{F}^{\text{Rig}}_{t \to s}$ in the target view based on the following relation:
	\begin{equation}
		\label{p_s}
		\tilde{\boldsymbol{p}}_s = \tilde{\boldsymbol{p}}_t + \begin{pmatrix}
			\boldsymbol{f}^{\text{Rig}}_{t\to s}\\ 0
		\end{pmatrix} = \frac{1}{\hat{z}_s}\boldsymbol{K}\left(
		\hat{z}_t \hat{\boldsymbol{R}}\boldsymbol{K}^{-1} \tilde{\boldsymbol{p}}_t + \hat{\boldsymbol{t}} \right),
	\end{equation}
	where 
	$\boldsymbol{f}^{\text{Rig}}_{t\to s}=\boldsymbol{F}^{\text{Rig}}_{t \to s}(\boldsymbol{p}_t)$ denotes the rigid flow at pixel $\boldsymbol{p}_t$ and is expected to equal the optical flow $\boldsymbol{f}^{\text{Opt}}_{t\to s}=\boldsymbol{F}^{\text{Opt}}_{t \to s}(\boldsymbol{p}_t)=\boldsymbol{p}_s - \boldsymbol{p}_t$ in static regions.
	$\boldsymbol{I}_s$ is then warped into the target view based on the rigid flow map, resulting in a synthesized image $\boldsymbol{I}'_t$ expressed as follows \cite{zhou2017unsupervised}:
	\begin{equation}
		\boldsymbol{I}'_{t} = \mathcal{W}\left(\boldsymbol{I}_{s}, \boldsymbol{F}^{\text{Rig}}_{t \to s}\right), 
	\end{equation}
	where $\mathcal{W}(\cdot)$ denotes the image warping function. To quantify the appearance discrepancy between $\boldsymbol{I}'_t$ and $\boldsymbol{I}_t$, the following photometric reprojection loss is computed: 
	\begin{equation}
		\mathcal{L}_\text{pho}=\alpha \frac{1-\textrm{SSIM}(\boldsymbol{I}'_t,\boldsymbol{I}_t)}{2} +(1-\alpha )\left \| \boldsymbol{I}' _t-\boldsymbol{I}_t \right \|_1,
		\label{eq.photometric_loss}
	\end{equation}
	where $\textrm{SSIM}$ denotes the pixel-wise structural similarity index operation \cite{wang2004image}, and the weight $\alpha$ is empirically set to 0.85, following the study \cite{godard2019digging}. \eqref{eq.photometric_loss} serves as the supervisory signal for the training of both DepthNet and PoseNet \cite{bian2021unsupervised}.

	\section{Discriminative Treatment of Motion Components}
	\label{sect.method1}
	Ego-motion estimation plays a role equally important to that of depth estimation, as the supervisory signal is jointly determined by the transformation $\boldsymbol{T}$ predicted by PoseNet and the depth map $\boldsymbol{D}_t$ generated by DepthNet. Unfortunately, ego-motion estimation has long been regarded as an auxiliary task \cite{godard2019digging,zhang2023lite}, with limited attention given to thoroughly analyzing the outputs produced by PoseNet.
	As discussed in Sect.~\ref{sect.introduction}, $\boldsymbol{T}$ is a mixture of three types of transformations as follows:
	\begin{equation}
		\label{eq.transformation}
		\boldsymbol{T}=
		\begin{pmatrix} \boldsymbol{R} & \boldsymbol{t} \\ \boldsymbol{0}^{\top} & 1 \end{pmatrix}=
		\underbrace{\begin{pmatrix} \boldsymbol{I} & {\boldsymbol{t}^\text{Rad}} \\ \boldsymbol{0}^{\top} & 1 \end{pmatrix}}_{\boldsymbol{T}^\text{Rad}}
		\underbrace{\begin{pmatrix} \boldsymbol{I} & {\boldsymbol{t}^\text{Tan}} \\ \boldsymbol{0}^\top & 1 \end{pmatrix}}_{\boldsymbol{T}^\text{Tan}}
		\underbrace{\begin{pmatrix} \boldsymbol{R} & \boldsymbol{0} \\ \boldsymbol{0}^{\top} & 1 \end{pmatrix}}_{\boldsymbol{T}^\text{Rot}},
	\end{equation}
	where $\boldsymbol{T}^\text{Rot}$, $\boldsymbol{T}^\text{Tan}$, and $\boldsymbol{T}^\text{Rad}$ represent pure rotation, pure tangential translation, and pure radial translation transformations, respectively. The prior art \cite{bian2021auto} claims that $\boldsymbol{T}^\text{Rot}$ is irrelevant to the training of DepthNet, as in the pure rotation case, one image can be warped to the other using $\boldsymbol{T}^\text{Rot}$ alone, without the need for depth information \cite{hartley2003multiple}. More importantly, even slight rotational errors in $\boldsymbol{T}^\text{Rot}$ can be mistakenly interpreted as translational motions, leading to erroneous estimations of $\boldsymbol{T}^\text{Tan}$ and $\boldsymbol{T}^\text{Rad}$. The resulting supervisory signals can, in turn, mislead the training of DepthNet. Therefore, the authors proposed to exclude rotational motions from DepthNet training within the joint learning framework. Specifically, the original target and source images, $\boldsymbol{I}_t$ and $\boldsymbol{I}_s$, are respectively warped based on $\hatsup{\boldsymbol{T}}{Rot}$, which is estimated using either the five-point algorithm \cite{nister2004efficient} or an auxiliary network \cite{he2016deep}, to align the orientations of source and target cameras. The warped image pair is then fed into the joint learning framework, enabling both PoseNet and DepthNet to be trained under the assumption that only translational motions are involved. Nonetheless, rotation estimation performed by an independent algorithm or network remains decoupled from the original joint learning framework, which introduces considerable computational and storage overhead without directly benefiting PoseNet training. 
	Moreover, rotational errors introduced by decoupled algorithms are no longer considered for minimization during training, allowing them to propagate through the pipeline and degrade overall framework performance.
	More critically, all existing methods overlook the distinct regularities inherent in the two types of translational motions, which are closely tied to depth estimation.
	
	Therefore, this study investigates the discriminative treatment of three distinct motion components to simultaneously improve both depth and pose estimation performance within the joint learning framework. Specifically, when the ego-motion corresponds to a pure rotation transformation $\boldsymbol{T}^\text{Rot}$, 
	the pixel correspondences $\boldsymbol{p}_t$ and $\boldsymbol{p}_s$ are expected to satisfy the following relation:
	\begin{equation}
		\label{eq.rotational_correspondence}
		\tilde{\boldsymbol{p}}_s = \frac{z_t}{z_s}\underbrace{\boldsymbol{K} \boldsymbol{R}
			\boldsymbol{K}^{-1}}_{\boldsymbol{H}} \tilde{\boldsymbol{p}}_t 
		\simeq \underbrace{
			\begin{pmatrix}
				\boldsymbol{h}_1, \boldsymbol{h}_2, \boldsymbol{h}_3
			\end{pmatrix}^\top}
		_{\boldsymbol{H}} \tilde{\boldsymbol{p}}_t,
	\end{equation}
	where $\boldsymbol{H}$ denotes the homography matrix \cite{hartley2003multiple} with $\boldsymbol{h}_i^{\top}=(h_{i1}, h_{i2}, h_{i3})$ representing the $i$-th row vector, and the symbol $\simeq$ indicates that the two vectors are equal up to a scale factor. Rearranging \eqref{eq.rotational_correspondence} yields
	\begin{equation}
		\label{eq.rotational_correspondence_h}
		\boldsymbol{p}_s = \left( \frac{\boldsymbol{h}_1^{\top}\tilde{\boldsymbol{p}}_t}{\boldsymbol{h}_3^{\top}\tilde{\boldsymbol{p}}_t},\frac{\boldsymbol{h}_2^{\top}\tilde{\boldsymbol{p}}_t}{\boldsymbol{h}_3^{\top}\tilde{\boldsymbol{p}}_t}  \right)^{\top}
	\end{equation}
	and 
	\begin{equation}
		\label{eq.rotational_flow}
		\boldsymbol{f}_{t \to s}^{\text{Rig}} = \frac{1}{\boldsymbol{h}_3^{\top}\tilde{\boldsymbol{p}}_t}\begin{pmatrix}
			\tilde{\boldsymbol{p}}_t^{\top}\left(-\boldsymbol{h}_3\boldsymbol{i}_1^{\top}\right)\tilde{\boldsymbol{p}}_t + \boldsymbol{h}_1^{\top}\tilde{\boldsymbol{p}}_t\\
			\tilde{\boldsymbol{p}}_t^{\top}\left(-\boldsymbol{h}_3\boldsymbol{i}_2^{\top}\right)\tilde{\boldsymbol{p}}_t + \boldsymbol{h}_2^{\top}\tilde{\boldsymbol{p}}_t\end{pmatrix},
	\end{equation}
	where $\boldsymbol{i}_k \in \mathbb{R}^3$ denotes the $k$-th column vector of an $3\times3$ identity matrix. \eqref{eq.rotational_flow} indicates that, when only rotation is involved, the rigid flow becomes a nonlinear function of the pixel coordinates $\boldsymbol{p}_t$. This nonlinearity arises from two sources: (1) the quadratic terms $\tilde{\boldsymbol{p}}_t^{\top}(-\boldsymbol{h}_3\boldsymbol{i}_1^{\top})\tilde{\boldsymbol{p}}_t=h_{31}u_t^2+h_{32}u_tv_t+h_{33}u_t$ and $\tilde{\boldsymbol{p}}_t^{\top}(-\boldsymbol{h}_3\boldsymbol{i}_2^{\top})\tilde{\boldsymbol{p}}_t=h_{31}u_tv_t+h_{32}v_t^2+h_{33}v_t$, as well as (2) the perspective division term $1/(\boldsymbol{h}_3^{\top}\tilde{\boldsymbol{p}}_t)=1/(h_{31}u_t+h_{32}v_t+h_{33})$. As a result, in such cases, the rigid flow is determined solely by the rotation and the intrinsic matrix, and can vary significantly across the entire image, giving rise to a highly irregular rigid flow map\footnote{Although rotational flows also comprises three relatively regular components \cite{bowen2022dimensions,poggi2025flowseek}, they cannot be obtained practically by decomposing the original flows. This is because the translational component cannot be eliminated prior to the rotational ones due to the non-commutative nature of rigid-body transformations, as shown in \eqref{eq.transformation}.}, as shown in Fig. \ref{fig.motivation}(c). 
	
	\begin{figure}[!t]
		\centering
		\includegraphics[width=0.49\textwidth]{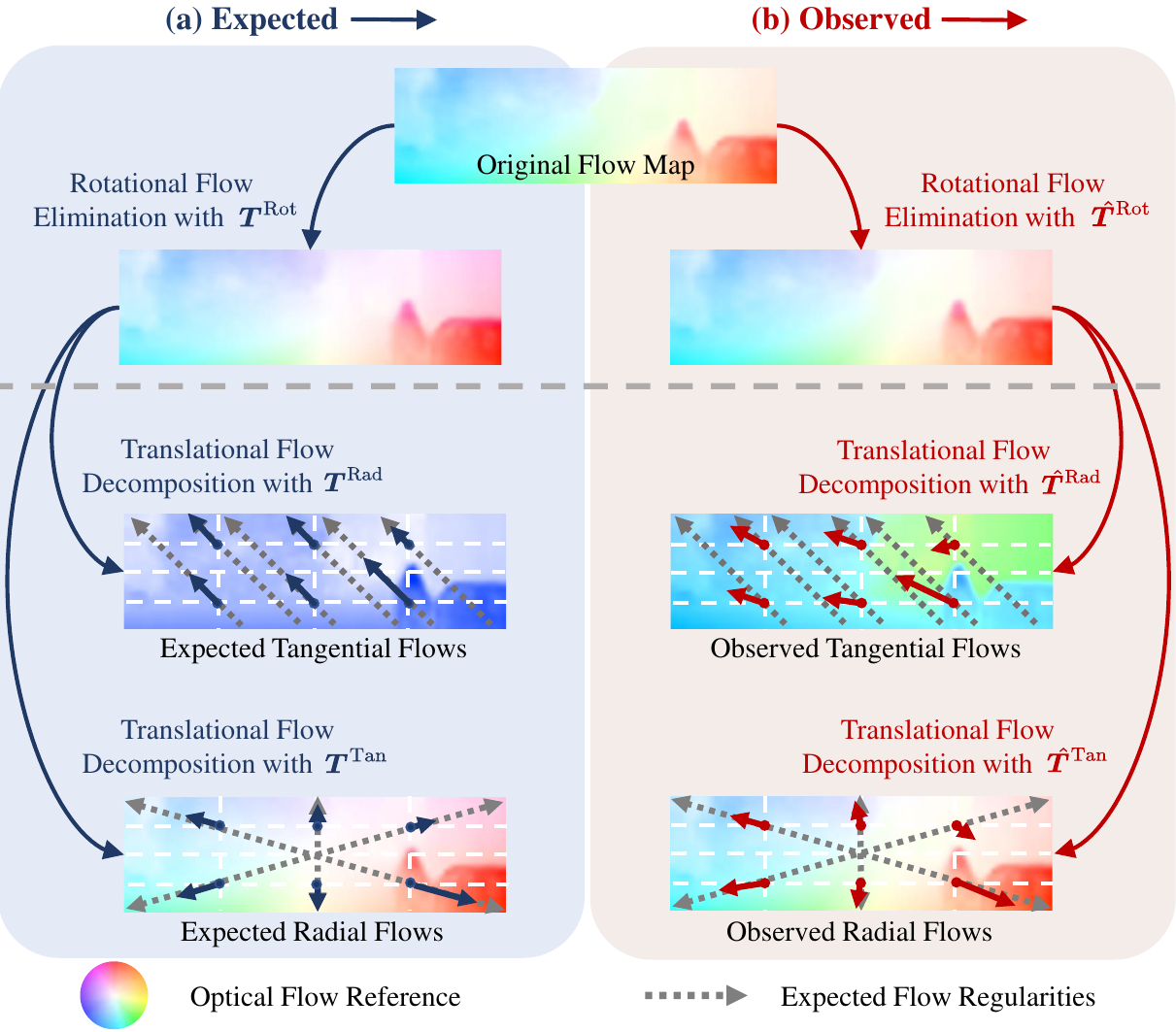}
		\caption{The discriminative treatment of motion components and the resulting flow decomposition processes.}
		\label{fig.decomposition}
	\end{figure}
	
	Therefore, instead of relying on an independent algorithm or auxiliary network to estimate rotation for image warping, we directly utilize $\hatsup{\boldsymbol{T}}{Rot}$ provided by the PoseNet to eliminate the irregular rotational flows from the original optical flow map. Ideally, the resulting rigid flow map retains only the two types of translational flows, induced by $\boldsymbol{T}^\text{Tan}$ and $\boldsymbol{T}^\text{Rad}$, which are directly relevant to depth estimation, as illustrated in Fig.~\ref{fig.decomposition}(a).
	
	When the ego-motion corresponds to a pure tangential translation transformation $\boldsymbol{T}^{\text{Tan}}$, the pixel correspondences $\boldsymbol{p}_t$ and $\boldsymbol{p}_s$ are expected to satisfy the following relation:
	\begin{equation}
		\label{eq.tangential_correspondence}
		\begin{aligned}
			\tilde{\boldsymbol{p}}_s
			\simeq \boldsymbol{K} \left(
			z_t \boldsymbol{K}^{-1} \tilde{\boldsymbol{p}}_t + \boldsymbol{t}^{\text{Tan}} \right).
		\end{aligned}
	\end{equation}
	Substituting \eqref{eq.intrinsic_matrix} into \eqref{eq.tangential_correspondence} yields
	\begin{equation}
		\label{eq.tangential_correspondence_2}
		\begin{aligned}
			{\boldsymbol{p}}_s = \left( \frac{z_tu_t+f_ut_x}{z_t}, \frac{z_tv_t+f_vt_y}{z_t}\right)^{\top}
		\end{aligned}
	\end{equation}
	and
	\begin{equation}
		\boldsymbol{f}_{t \to s}^{\text{Rig}} = \frac{1} {z_t} 
		\begin{pmatrix}f_ut_x\\f_vt_y\end{pmatrix}, 
		\label{eq.tangential_flow}
	\end{equation}
	which indicates that the tangential translational flows across the image are oriented in the direction of $(f_ut_x, f_vt_y)^{\top}$ and vary solely with depth, as illustrated in Fig.~\ref{fig.decomposition}(a).
	Similarly, when the ego-motion corresponds to a pure radial translation transformation $\boldsymbol{T}^{\text{Rad}}$, the pixel correspondences $\boldsymbol{p}_t$ and $\boldsymbol{p}_s$ are expected to satisfy the following relation:
	\begin{equation}
		\label{eq.radial_correspondence}
		\begin{aligned}
			\tilde{\boldsymbol{p}}_s
			\simeq \boldsymbol{K} \left(
			z_t \boldsymbol{K}^{-1} \tilde{\boldsymbol{p}}_t + \boldsymbol{t}^{\text{Rad}} \right). 
		\end{aligned}
	\end{equation}
	Substituting \eqref{eq.intrinsic_matrix} into \eqref{eq.radial_correspondence} yields
	\begin{equation}
		\label{eq.radial_correspondence_2}
		\begin{aligned}
			{\boldsymbol{p}}_s = \left ( \frac{z_tu_t+u_0t_z}{z_t+t_z}, \frac{z_tv_t+v_0t_z}{z_t+t_z} \right )^{\top}
		\end{aligned}
	\end{equation}
	and
	\begin{equation}
		\boldsymbol{f}_{t \to s}^{\text{Rig}}
		= \frac{-t_z }{z_t+t_z}\begin{pmatrix}u_t-u_0\\v_t-v_0\end{pmatrix},
		\label{eq.radial_flow}
	\end{equation}
	which indicates that radial translational flows across the image are consistently oriented toward or away from the principal point $\boldsymbol{p}_0$, as illustrated in Fig.~\ref{fig.decomposition}(a). 
	Consequently, these flows are relevant not only with depth but also with pixel coordinates, with their magnitudes diminishing near $\boldsymbol{p}_0$ due to the reduced distance between $\boldsymbol{p}_t$ and $\boldsymbol{p}_0$. However, when both $\boldsymbol{T}^{\text{Tan}}$ and $\boldsymbol{T}^{\text{Rad}}$ are present, the resulting rigid flows can be expressed as follows:
	\begin{equation}
		\boldsymbol{f}_{t \to s}^{\text{Rig}} = \frac{1} {z_t+t_z} 
		\begin{pmatrix}f_ut_x-(u_t-u_0)t_z\\f_vt_y-(v_t-v_0)t_z\end{pmatrix},
	\end{equation}
	which obscures the distinct regularities inherent in both types of flows, thereby hindering the framework from discriminately backpropagating their corresponding supervisory signals to the network outputs.
	Specifically, the gradients propagated through $\boldsymbol{f}_{t \to s}^{\text{Rig}}$ are given by:
	\begin{align}
		\label{eq.grad_z}
		&\dfrac{ \partial\boldsymbol{f}_{t \to s}^{\text{Rig}}}{ \partial z_t} =
		-\dfrac{1}{(z_t+t_z)^2}\begin{pmatrix}
			f_ut_x-(u_t-u_0)t_z \\
			f_vt_y-(v_t-v_0)t_z
		\end{pmatrix}, \\
		\label{eq.grad_tx}
		&\dfrac{ \partial\boldsymbol{f}_{t \to s}^{\text{Rig}}}{ \partial t_x} =
		\dfrac{1}{z_t+t_z}\begin{pmatrix} f_u \\ 0 \end{pmatrix}, \\
		\label{eq.grad_ty}
		&\dfrac{ \partial\boldsymbol{f}_{t \to s}^{\text{Rig}}}{ \partial t_y} =
		\dfrac{1}{z_t+t_z}\begin{pmatrix} 0 \\ f_v \end{pmatrix}, \\
		\label{eq.grad_tz}
		&\frac{ \partial\boldsymbol{f}_{t \to s}^{\text{Rig}}}{ \partial t_z} =
		-\frac{1}{(z_t+t_z)^2}\begin{pmatrix}
			(u_t-u_0) z_t + f_u t_x \\
			(v_t-v_0) z_t + f_v t_y
		\end{pmatrix}.
	\end{align}
	\eqref{eq.grad_z} indicates that $z_t$ inherently receives smaller gradients near the principal point, which hinders consistent optimization across the image.
	Moreover, \eqref{eq.grad_tx} and \eqref{eq.grad_ty} show that the gradient of tangential translation components are influenced by the radial one, while \eqref{eq.grad_tz} reveals the opposite influence. As a result, errors in one component can propagate to and impair the gradients for the others, ultimately hindering the convergence of PoseNet training.
	
	To address these issues, after eliminating the irregular rotational flows, we further utilize \(\hatsup{\boldsymbol{T}}{Tan}\) and \(\hatsup{\boldsymbol{T}}{Rad}\) to decompose the remaining translational flows into the components that are geometrically regular but different with respect to depth, as illustrated in Fig.~\ref{fig.decomposition}(a). The decomposed tangential and radial flows are then used to incorporate additional geometric constraints (detailed in Sect.~\ref{sect.method2_constraint}) into the joint learning framework, enabling independent refinement of depth and each translational component.
	
	\begin{figure}[!t]
		\centering
		\includegraphics[width=0.49\textwidth]{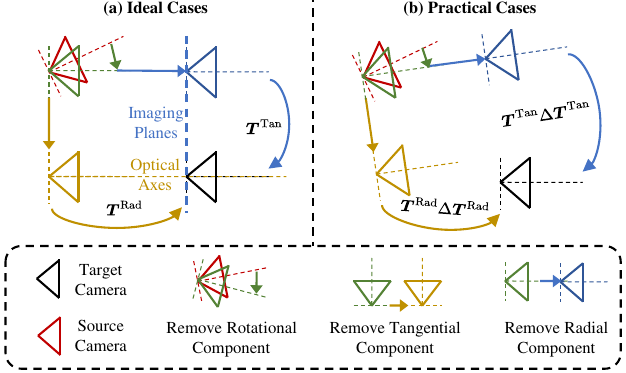}
		\caption{Ego-motion transformation decomposition: (a) Ideally, ego-motion can be decomposed into pure tangential and radial translation components; (b) In practice, errors in the PoseNet predictions introduce undesired deviations to the decomposed components.}
		\label{fig.residual}
	\end{figure}
	
	The above flow decomposition processes are performed by removing the rotational component from the full transformation $\boldsymbol{T}$, followed by the simultaneous and independent removal of the radial and tangential components. The two ideal decomposition processes, as illustrated in Fig.~\ref{fig.residual}(a), can be formulated by rearranging \eqref{eq.transformation} as follows:
	\begin{equation}
		\begin{cases}
			\begin{aligned}  \boldsymbol{T}^{\text{Rad}}=\boldsymbol{T}\left(\boldsymbol{T}^{\text{Rot}}\right)^{-1}\left(\boldsymbol{T}^{\text{Tan}}\right)^{-1} \\
				\boldsymbol{T}^{\text{Tan}}=\boldsymbol{T}\left(\boldsymbol{T}^{\text{Rot}}\right)^{-1}\left(\boldsymbol{T}^{\text{Rad}}\right)^{-1}
			\end{aligned}.
		\end{cases}
		\label{fig.idealtransformation}
	\end{equation}
	However, in practice, errors in the ego-motion estimated by PoseNet are inevitable, which can introduce deviations $\Delta\boldsymbol{T}^{\text{Rad}}$ and $\Delta\boldsymbol{T}^{\text{Tan}}$ into the above two decomposition processes, respectively, resulting in the following expression:
	\begin{equation}
		\begin{cases}
			\begin{aligned}
				\boldsymbol{T}^{\text{Rad}}\Delta\boldsymbol{T}^{\text{Rad}}=\boldsymbol{T}\left(\hatsup{\boldsymbol{T}}{Rot}\right)^{-1}\left(\hatsup{\boldsymbol{T}}{Tan}\right)^{-1} \\
				\boldsymbol{T}^{\text{Tan}}\Delta\boldsymbol{T}^{\text{Tan}}=\boldsymbol{T}\left(\hatsup{\boldsymbol{T}}{Rot}\right)^{-1}\left(\hatsup{\boldsymbol{T}}{Rad}\right)^{-1}
			\end{aligned}.
		\end{cases}
		\label{fig.practicaltransformation}
	\end{equation}
	Such deviations, shown in Fig.~\ref{fig.residual}(b), can be derived from \eqref{fig.practicaltransformation} and expressed as follows:
	\begin{equation}
		\begin{cases}
			\begin{aligned}
				\label{eq.residual_rot}
				\Delta\boldsymbol{T}^{\text{Rad}}
				&=\begin{pmatrix} 
					\Delta\boldsymbol{R}^{-1} & (\boldsymbol{I}-\Delta\boldsymbol{R}^{-1})\boldsymbol{t}^{\text{Tan}}-\Delta\boldsymbol{R}^{-1}\Delta \boldsymbol{t}^{\text{Tan}} \\ 
					\boldsymbol{0}^{\top} & 1
				\end{pmatrix} \\
				\Delta\boldsymbol{T}^{\text{Tan}}
				&=\begin{pmatrix}
					\Delta\boldsymbol{R}^{-1} & (\boldsymbol{I}-\Delta\boldsymbol{R}^{-1})\boldsymbol{t}^{\text{Rad}}-\Delta\boldsymbol{R}^{-1}\Delta \boldsymbol{t}^{\text{Rad}} \\ 
					\boldsymbol{0}^{\top} & 1
				\end{pmatrix}
			\end{aligned}.
		\end{cases}
	\end{equation}
	where \(\Delta \boldsymbol{t}^{\text{Tan}}=\hatsup{\boldsymbol{t}}{Tan}-\boldsymbol{t}^{\text{Tan}}\) and 
	\(\Delta \boldsymbol{t}^{\text{Rad}}=\hatsup{\boldsymbol{t}}{Rad}-\boldsymbol{t}^{\text{Rad}}\) represent the translational deviations in \(\hatsup{\boldsymbol{T}}{Tan}\) and \(\hatsup{\boldsymbol{T}}{Rad}\), respectively, while $\Delta\boldsymbol{R} = \hat{\boldsymbol{R}}\boldsymbol{R}^{-1}$ denotes the rotational deviation.
	As shown in \eqref{eq.residual_rot}, $\Delta\boldsymbol{T}^{\text{Rad}}$ and $\Delta\boldsymbol{T}^{\text{Tan}}$ introduce not only the rotational deviation $\Delta\boldsymbol{R}$ but also the other type of translational component into $\boldsymbol{T}^{\text{Rad}}$ and $\boldsymbol{T}^{\text{Tan}}$, respectively. Consequently, as illustrated in Fig.~\ref{fig.decomposition}(b), the resulting radial and tangential translational flows deviate from their expected ones, thereby impairing the effectiveness of incorporating additional geometric constraints derived from optical flow decomposition.
	
	To address the aforementioned practical challenges, we leverage two geometric constraints (detailed in Sect.~\ref {sect.method2_constraint}) to jointly minimize the deviations between the decomposed translational flows and the expected ones. These two loss functions collaboratively guide the estimation of $\hatsup{\boldsymbol{T}}{Rot}$, while simultaneously providing independent supervisory signals to $\hatsup{\boldsymbol{T}}{Tan}$ and $\hatsup{\boldsymbol{T}}{Rad}$.
	
	In widely used driving datasets, such as KITTI \cite{geiger2012we} and nuScenes \cite{caesar2020nuscenes}, vehicles predominantly move forward, with only gradual rotations occurring during turns. As a result, these datasets present limited challenges for accurate rotation estimation. Nevertheless, since $\boldsymbol{T}^{\text{Rad}}$ typically dominates over $\boldsymbol{T}^{\text{Tan}}$, even a minor rotational deviation $\Delta\boldsymbol{R}$ can induce a noticeable undesired deviation $(\boldsymbol{I}-\Delta\boldsymbol{R})\boldsymbol{t}^{\text{Rad}}$ in $\boldsymbol{t}^{\text{Tan}}$. To further demonstrate the robustness of the proposed method under more challenging conditions involving substantial residual rotations $\Delta\boldsymbol{R}$, we create a new dataset (detailed in Sect.~\ref{Sect.datasets}) containing sequences with pronounced rotational motion and frequent camera shake.
	
	\section{DiMoDE}
	\label{sect.method2}
	
	\begin{figure*}[t!]
		\centering
		\includegraphics[width=0.999 \textwidth]{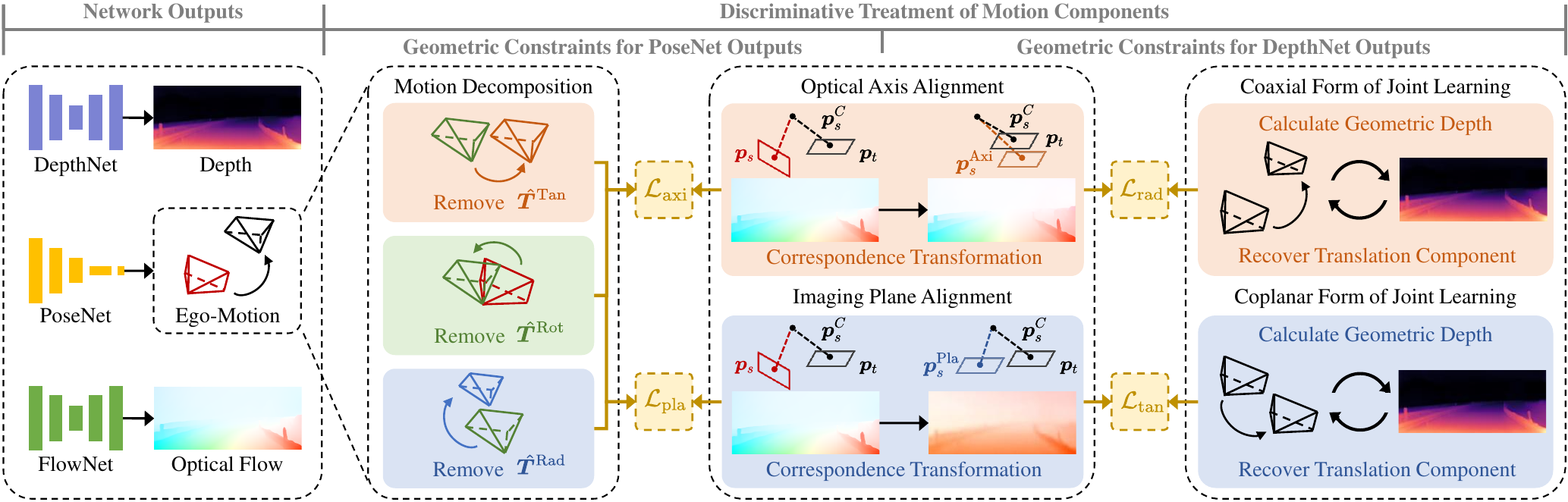}
		\caption{An overview of the DiMoDE framework. Centered on the core idea of discriminative motion component treatment, depth (from DepthNet) and ego-motion (from PoseNet) are utilized to perform optical axis and imaging plane alignments. FlowNet generates dense correspondences, which are transformed during the alignment processes. The transformed flows are ultimately leveraged to incorporate two sets of geometric constraints into the unsupervised joint learning framework, thereby simultaneously improving both depth and pose estimation performance.}
		\label{fig.overall_architect}
	\end{figure*}

	As shown in Fig.~\ref{fig.residual}(a), removing the rotational and tangential components from the full transformation is equivalent to aligning the optical axes of the source and target cameras, while removing the rotational and radial components is equivalent to aligning their imaging planes. To realize the proposed discriminative treatment of motion components in practice, we incorporate such alignment processes into the joint learning framework, thereby introducing explicit geometric constraints that refine the outputs of both networks.
	
	\subsection{Framework Overview}
	\label{sect.method2_overview}
	An overview of the proposed DiMoDE framework is presented in Fig.~\ref{fig.overall_architect}. The ego-motion transformation $\hat{\boldsymbol{T}}$, estimated by PoseNet, is decomposed into three components: \(\hatsup{\boldsymbol{T}}{Rot}\), \(\hatsup{\boldsymbol{T}}{Tan}\) and \(\hatsup{\boldsymbol{T}}{Rad}\). First, \(\hatsup{\boldsymbol{T}}{Rot}\) is applied to the source camera to align its orientation with that of the target camera. Subsequently, \(\hatsup{\boldsymbol{T}}{Tan}\) and \(\hatsup{\boldsymbol{T}}{Rad}\) are used to align the optical axes and imaging planes, respectively. In accordance with these two alignment processes, the original optical flows provided by FlowNet are transformed to those from the target image to the aligned source camera image. 
	To refine each component of the PoseNet output, the first set of geometric constraints is imposed by minimizing the deviations between the transformed optical flows and the expected pure radial and tangential ones. Once the PoseNet predictions achieve sufficient accuracy, ensuring proper alignment of optical axes and imaging planes, each translation component is incorporated individually to further refine the DepthNet outputs. In such cases, the depth and each translation component can be mutually determined through the decomposed flows, thereby forming constraint cycles that provide robust supervisory signals for DepthNet training under adverse conditions.
	
	\subsection{Optical Axis and Imaging Plane Alignments}
	\label{sect.method2_constraint}
	
	To integrate the alignment processes and their corresponding geometric constraints into the joint learning framework, a straightforward approach, adopted in the previous study \cite{bian2021auto}, is to reconstruct the warped image once after removing each motion component and then establish dense correspondences from the warped image pair.
	However, repeated image warping incurs substantial computational overhead and often introduces holes and artifacts in the reconstructed images \cite{zhou2017unsupervised}, which undermine the reliability of correspondence generation and ultimately weaken the effectiveness of the imposed geometric constraints.
	To address this issue, we adopt a correspondence transformation strategy that directly transforms the original optical flow $\boldsymbol{f}_{t \to s}^{\text{Opt}}$ into the optical flows resulting from optical axis and imaging plane alignment, denoted as $\boldsymbol{f}_{t \to s}^{\text{Axi}}$ and $\boldsymbol{f}_{t \to s}^{\text{Pla}}$, respectively.
	Specifically, pixel correspondences $\boldsymbol{p}_s$ and $\boldsymbol{p}_t$ are generated once using the target-to-source optical flow map $\boldsymbol{F}_{t \to s}^{\text{Opt}}$ predicted by FlowNet \cite{teed2020raft}.
	To backproject $\boldsymbol{p}_s$ into 3D camera coordinates, the source-view depth map $\boldsymbol{D}_s$ is warped into the target view based on the optical flow map, producing the following pixel-aligned depth map \cite{bian2021unsupervised, shao2024monodiffusion}:
	\begin{equation}
		\boldsymbol{D}_{s\to t} = \mathcal{W}\left(\boldsymbol{D}_{s}, \boldsymbol{F}_{t\to s}^{\text{Opt}}\right). 
	\end{equation}
	$\boldsymbol{p}^C_s$, the 3D camera coordinates of $\boldsymbol{p}_s$, can be computed using the following expression:
	\begin{equation}
		\boldsymbol{p}^C_{s} = \boldsymbol{D}_{s \to t}(\boldsymbol{p}_t) \boldsymbol{K}^{-1} \boldsymbol{p}_{s}.
	\end{equation}
	Transforming $\boldsymbol{p}^C_{s}$ through the alignment processes and projecting it back into pixel coordinates yields the following expressions:
	\begin{equation}
		\label{pplanar}
		\begin{cases}
			\begin{aligned}
				\boldsymbol{p}_{s}^{\text{Pla}} \simeq \boldsymbol{K}\left(\hat{\boldsymbol{R}}^{-1}\boldsymbol{p}^C_{s}  - \hatsup{\boldsymbol{t}}{Rad}\right) \\
				\boldsymbol{p}_{s}^{\text{Axi}} \simeq \boldsymbol{K}\left(\hat{\boldsymbol{R}}^{-1}\boldsymbol{p}^C_{s}  - \hatsup{\boldsymbol{t}}{Tan}\right)
			\end{aligned}.
		\end{cases}
	\end{equation}
	Finally, the optical flows $\boldsymbol{f}_{t \to s}^{\text{Pla}}$ and $\boldsymbol{f}_{t \to s}^{\text{Axi}}$ are derived as follows:
	\begin{equation}
		\label{disflow}
		\begin{cases}
			\begin{aligned}
				\boldsymbol{f}_{t \to s}^{\text{Pla}} = &\boldsymbol{p}_{s}^{\text{Pla}}-\boldsymbol{p}_{t}= \displaystyle\frac{\boldsymbol{K}\left(\hat{\boldsymbol{R}}^{-1}\boldsymbol{p}^C_{s}  - \hatsup{\boldsymbol{t}}{Rad}\right)}{\boldsymbol{i}_3^{\top}\left(\hat{\boldsymbol{R}}^{-1}\boldsymbol{p}^C_{s}  - \hatsup{\boldsymbol{t}}{Rad}\right)} -\boldsymbol{p}_{t} \\
				\boldsymbol{f}_{t \to s}^{\text{Axi}} = &\boldsymbol{p}_{s}^{\text{Axi}}-\boldsymbol{p}_{t}= \displaystyle\frac{\boldsymbol{K}\left(\hat{\boldsymbol{R}}^{-1}\boldsymbol{p}^C_{s}  - \hatsup{\boldsymbol{t}}{Tan}\right)}{\boldsymbol{i}_3^{\top}\left(\hat{\boldsymbol{R}}^{-1}\boldsymbol{p}^C_{s}  - \hatsup{\boldsymbol{t}}{Tan}\right)} -\boldsymbol{p}_t
			\end{aligned}.
		\end{cases}
	\end{equation}
	\subsection{Geometric Constraints from Motion Components}
	\label{sect.method2_constraint}
	
	As discussed in Sect.~\ref{sect.method1}, $\boldsymbol{f}_{t \to s}^{\text{Pla}}$ and $\boldsymbol{f}_{t \to s}^{\text{Axi}}$ are expected to approximate the tangential and radial translational flows.
	The former should exhibit a uniform direction across the image, while the latter should point toward or away from the principal point at each pixel. Based on \eqref{eq.tangential_flow} and \eqref{eq.radial_flow}, these properties enable the formulation of two geometric constraints: (1) all $\boldsymbol{f}_{t \to s}^{\text{Pla}}$ vectors form a consistent angle with a global reference vector $\boldsymbol{e}_1$, such that $\frac{\boldsymbol{e}_1^{\top}\boldsymbol{f}_{t \to s}^{\text{Pla}}}{\left \| \boldsymbol{f}_{t \to s}^{\text{Pla}} \right \|_2}$ remains constant across the image, where $\boldsymbol{e}_k \in \mathbb{R}^2$ denotes the $k$-th column vector of the $2\times2$ identity matrix; (2) each $\boldsymbol{f}_{t \to s}^{\text{Axi}}$ vector is expected to be parallel to the direction from its pixel $\boldsymbol{p}_t$ to the principal point $\boldsymbol{p}_0$, such that $\frac{\boldsymbol{f}_{t \to s}^{\text{Axi}}}{\left \| \boldsymbol{f}_{t \to s}^{\text{Axi}} \right \|_2 } = \frac{\boldsymbol{p}_t-\boldsymbol{p}_0}{\left \| \boldsymbol{p}_t-\boldsymbol{p}_0 \right \|_2 }$.
	In practice, these two geometric constraints are imposed by incorporating the following loss functions into the joint learning framework:
	\begin{equation}
		\label{eq.lp}
		\mathcal{L}_\text{pla} = \mathrm{Var} \left( \arccos\frac{ \boldsymbol{e}_1^\top \boldsymbol{f}_{t \to s}^{\text{Pla}} }{\| \boldsymbol{f}_{t \to s}^{\text{Pla}} \|_2}\right),
	\end{equation}
	and
	\begin{equation}
		\label{eq.la}
		\mathcal{L}_\text{axi} = \mathbb{E} \left[ \arccos\frac{ (\boldsymbol{f}_{t \to s}^{\text{Axi}})^\top \left(\boldsymbol{p}_t - \boldsymbol{p}_0\right) }{ \| \boldsymbol{f}_{t \to s}^{\text{Axi}} \|_2 \, \| \boldsymbol{p}_t - \boldsymbol{p}_0 \|_2 } \right].
	\end{equation}
	where the mean $\mathbb{E}[\cdot]$ and variance $\mathrm{Var}(\cdot)$ are computed over all valid pixels in static and non-occluded regions, as determined by the masking technique proposed in the study \cite{sun2023dynamo}.
	Continuously imposing these constraints through \eqref{eq.lp} and \eqref{eq.la} enables more accurate imaging plane and optical axis alignments, respectively. The radial translation (used exclusively for the imaging plane alignment) and the tangential translation (used solely for the optical axes alignment) are therefore optimized independently, while the rotation estimation is jointly constrained by both. The effectiveness of the imposed constraints and the complementary role of \eqref{eq.lp} and \eqref{eq.la} are validated through an ablation study detailed in Sect.~\ref{sect.abl}.
	
	Once the PoseNet predictions reach sufficient accuracy, the alignment processes allow the original joint depth and ego-motion learning from monocular video to be reformulated into two geometrically simplified forms: the coaxial form, analogous to learning from axial-motion image pairs, and the coplanar form, analogous to learning from stereo image pairs along the horizontal and vertical directions.
	In such forms, the depth at each pixel and the ego-motion satisfy the following relations, derived from \eqref{eq.tangential_flow} and \eqref{eq.radial_flow}:
	\begin{equation}
		\label{eq.constraint3}
		{\rho_x}:=\dfrac{z_t}{t_x} = \frac{f_x}{ \boldsymbol{e}_1^{\top}\boldsymbol{f}_{t \to s}^{\text{Pla}} }, \quad
		{\rho_y}:=\dfrac{z_t}{t_y} = \frac{f_y}{ \boldsymbol{e}_2^{\top}\boldsymbol{f}_{t \to s}^{\text{Pla}} },
	\end{equation}
	\begin{equation}
		\label{eq.constraint4}
		{\rho_z}:=\dfrac{z_t}{t_z} = -\frac{\left(\boldsymbol{p}_t -\boldsymbol{p}_0\right)^{\top} (\boldsymbol{f}_{t \to s}^{\text{Axi}} + \boldsymbol{p}_t -\boldsymbol{p}_0) }{\left(\boldsymbol{p}_t -\boldsymbol{p}_0\right)^{\top} \boldsymbol{f}_{t \to s}^{\text{Axi}}}, 
	\end{equation}
	where $\rho_x$, $\rho_y$, and $\rho_z$ are defined as the per-pixel ratios between depth and the corresponding translation components. These ratios are determined by the transformed optical flows, with $\rho_z$ additionally depending on the pixel coordinates.
	According to \eqref{eq.constraint3} and \eqref{eq.constraint4}, at each pixel $\boldsymbol{p}_t$, multiplying the respective translation components by their corresponding ratios yields three geometric depth values: $\rho_x\hat{t}_x$, $\rho_y\hat{t}_y$, and $\rho_z\hat{t}_z$. These values are expected to be consistent with the depth $\hat{z}_t$ predicted by DepthNet, thereby imposing three per-pixel geometric constraints to supervise DepthNet training.
	Moreover, the simplified geometric relationships enable closed-form inverse computation: given the estimated depth and three ratio maps, each translation component can be recovered as $\mathbb{E}[\hat{z}_t/\rho_x]$, $\mathbb{E}[\hat{z}_t/\rho_y]$, and $\mathbb{E}[\hat{z}_t/\rho_z]$. This eliminates the need for iterative optimization. The recovered translation components are also expected to be consistent with those estimated by PoseNet. Based on these constraints, we introduce the following two loss functions to the joint learning framework: 
	\begin{equation}
		\begin{aligned}
			\label{eq.lt}
			\mathcal{L}_\text{tan} = &\mathbb{E} \left[ \frac{ | \rho_x\hat{t}_x-\hat{z}_t |}{\hat{z}_t} \right ] + \left | \frac{\mathbb{E} \left[\hat{z}_t/\rho_{x} \right]-\hat{t}_x}{\hat{t}_x} \right |\\
			+ &\mathbb{E} \left[ \frac{ | \rho_y\hat{t}_y-\hat{z}_t |}{\hat{z}_t} \right ] + \left | \frac{\mathbb{E} \left[\hat{z}_t/\rho_{y} \right]-\hat{t}_y}{\hat{t}_y} \right |. 
		\end{aligned}
	\end{equation}
	and
	\begin{equation}
		\label{eq.lr}
		\mathcal{L}_\text{rad} = \mathbb{E} \left[ \frac{ | \rho_z\hat{t}_z-\hat{z}_t |}{\hat{z}_t} \right ] + \left | \frac{\mathbb{E} \left[\hat{z}_t/\rho_{z} \right]-\hat{t}_z}{\hat{t}_z} \right |,
	\end{equation}
	By eliminating the pixel-coordinate dependence of $\boldsymbol{f}_{t \to s}^{\text{Axi}}$ via \eqref{eq.constraint4}, the resulting $\rho_z$ becomes a function of depth along. Consequently, the first term in both \eqref{eq.lt} and \eqref{eq.lr} can provide consistent gradients with respect to per-pixel depth errors across the entire map. Furthermore, unlike existing methods that impose only pixel-level constraints into DepthNet training \cite{sun2023dynamo, sun2023sc, zhang2024dcpi}, the proposed method establishes two constraint cycles via \eqref{eq.lt} and \eqref{eq.lr}, within which the second terms in both \eqref{eq.lt} and \eqref{eq.lr}, along with the fourth term in \eqref{eq.lt}, measure the overall depth errors aggregated over all valid pixels. Consequently, in challenging scenarios, such as night-time or adverse weather conditions, where per-pixel supervisory signals may become unreliable, this constraint cycle can help prevent significant degradation in DepthNet performance. To validate its effectiveness, we conduct extensive comparative experiments and ablation studies on the nuScenes dataset \cite{caesar2020nuscenes} (see Sects.~\ref{Sect.depth_results} and \ref{sect.abl}), which contains numerous video sequences captured under the above-mentioned challenging conditions.
	
	In addition to the above newly proposed loss functions, we retain the minimum photometric reprojection loss $\mathcal{L}_\text{pho}$ adopted in the study \cite{godard2019digging}, as well as the local structure refinement loss $\mathcal{L}_\text{loc}$ employed in the study \cite{sun2023sc} to establish the baseline training framework.
	The overall loss function is formulated as:
	\begin{equation}
		\label{eq.overall_loss}
		\mathcal{L} = \lambda_1 (\mathcal{L}_\text{axi} + \mathcal{L}_\text{pla}) +  \lambda_2 (\mathcal{L}_\text{rad}+ \mathcal{L}_\text{tan}) + \lambda_3 \mathcal{L}_\text{loc}+ \mathcal{L}_\text{pho},
	\end{equation}
	where $\lambda_1$, $\lambda_2 $, and $\lambda_3$ are weighting coefficients that balance the contributions of the corresponding loss terms. The specific weight settings and the overall training strategy are detailed in Sect.~\ref{Sect.implement}, and the effect of varying weighting coefficients is further analyzed in Sect.~\ref{sect.abl}.

	\section{Experiments}
	\label{Sect.experiments}
	
	Sect.~\ref{Sect.implement} provides implementation details on the proposed DiMoDE framework and the adopted training settings. Sect.~\ref{Sect.datasets} introduces the six public datasets used for performance comparison, along with a newly created real-world dataset designed to evaluate model performance in complex scenarios. Evaluation metrics are detailed in Sect.~\ref{Sect.metrics}. Sect.~\ref{Sect.vo_results} and Sect.~\ref{Sect.depth_results} present comprehensive comparisons between our method and previous SoTA approaches in monocular visual odometry and depth estimation, respectively. Finally, Sect.~\ref{sect.abl} provides detailed ablation studies to evaluate the efficacy of each innovative design based on discriminative motion component treatment, and details hyperparameter selection.
	
	\subsection{Implementation Details}
	\label{Sect.implement}
	
	As DiMoDE is designed as a general framework rather than being tied to a specific network architecture, we evaluate its compatibility with a variety of existing depth and pose estimation networks in our experiments.
	For DepthNet, we adopt Lite-Mono \cite{zhang2023lite}, a lightweight CNN-Transformer hybrid network, as well as D-HRNet \cite{he2022ra}, a representative CNN-based network, to evaluate DiMoDE’s adaptability to different depth estimation network architectures. Given the higher computational cost of D-HRNet, the majority of our experiments are conducted using Lite-Mono.
	For PoseNet, we adopt a lightweight design based on a ResNet-18 \cite{he2016deep} encoder, following previous studies \cite{godard2019digging, bian2021unsupervised}. To further validate the effectiveness of DiMoDE, we also employ the PoseNet proposed in the study \cite{feng2024scipad}, which incorporates spatial clues into the pose estimation process. However, since the latter PoseNet \cite{feng2024scipad} is significantly more computationally intensive and does not support real-time inference, we primarily adopt the basic ResNet-based PoseNet \cite{godard2019digging, bian2021unsupervised} in our experiments.
	Finally, RAFT \cite{teed2020raft} is employed as the FlowNet in DiMoDE to generate dense correspondences.
	
	All networks are trained on an NVIDIA RTX 4090 GPU. To fully utilize GPU memory, the batch size is set to 8 for images with a resolution of 640$\times$192 pixels. For higher resolutions, such as 640$\times$384 and 512$\times$288, the batch size is reduced to 4 to accommodate memory constraints. Following the study \cite{godard2019digging}, we utilize a snippet of three consecutive video frames as a training sample. For data augmentation, random color jitter and horizontal flips are applied to the images during model training. To minimize loss functions, we use the AdamW \cite{loshchilov2017decoupled} optimizer with an initial learning rate of $10^{-4}$ and a weight decay of $10^{-2}$.
	
	The encoders of DepthNet and PoseNet are initialized with pretrained weights from the ImageNet database \cite{deng2009imagenet}, following the studies \cite{godard2019digging, zhang2023lite}. FlowNet is pretrained on the synthetic FlyingChairs dataset \cite{dosovitskiy2015flownet}, following the study \cite{teed2020raft}.
	In the early training stage, network predictions tend to be less reliable. Directly enforcing optical axis and imaging plane alignments under such conditions may introduce significant deviations and lead to unstable training.
	To mitigate this problem, we adopt a progressive training strategy. Specifically, during the first 5 epochs out of the total 30 (Stage 1), we jointly train the three networks in an unsupervised manner using only $\mathcal{L}_\text{pho}$ and $\mathcal{L}_\text{loc}$, where the weights $(\lambda_1, \lambda_2,\lambda_3)$ in \eqref{eq.overall_loss} are set to $(0,0,0.01)$. In the subsequent 10 epochs (Stage 2), we freeze FlowNet and incorporate the alignment processes to impose the first set of geometric constraints, namely $\mathcal{L}_\text{pla}$ and $\mathcal{L}_\text{axi}$, primarily to refine pose estimation. The weights $(\lambda_1, \lambda_2, \lambda_3)$ are updated to $(0.05,0,0.01)$. Finally, during the last 15 epochs (Stage 3), we introduce the second set of geometric constraints, namely $\mathcal{L}_\text{tan}$ and $\mathcal{L}_\text{rad}$, to further improve depth estimation performance, with the weights $(\lambda_1, \lambda_2,\lambda_3)$ set to $(0.05,0.1,0.01)$.
	
	\begin{figure}[!t]
		{
			\setlength{\abovecaptionskip}{5pt}
			\centering
			\includegraphics[width=0.485\textwidth]{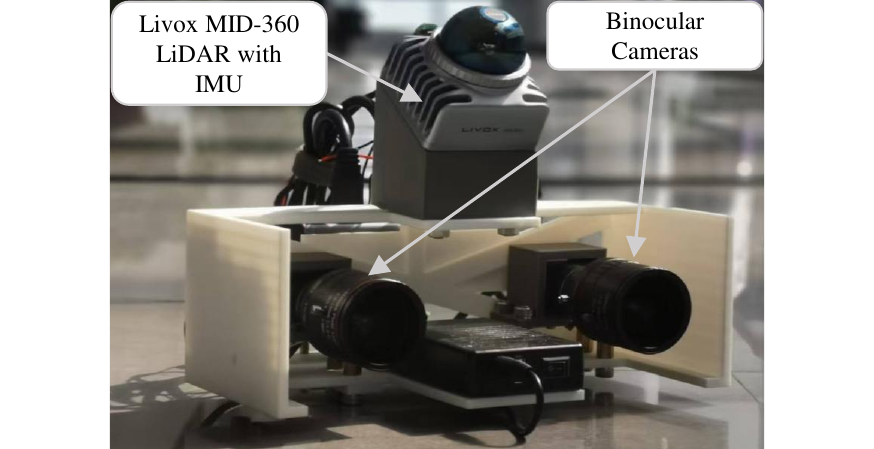}
			\caption{An illustration of our designed handheld setup equipped with calibrated and synchronized sensors for accurate real-world data collection.}
			\label{fig.handheld_setup}
		}
	\end{figure}
	
	\subsection{Datasets}
	\label{Sect.datasets}
	The performance of our method on visual odometry is evaluated using both the KITTI Odometry dataset \cite{geiger2013vision} and our newly created real-world MIAS-Odom dataset.
	The KITTI Odometry dataset contains 22 video sequences in a driving scenario, with ground-truth trajectories available for Seqs. 00-10. 
	Following prior works \cite{godard2019digging, bian2021unsupervised}, we use Seqs. 00–08 for unsupervised framework training. Seqs. 09-10 serve as the standard test set, while Seqs. 11–21 are additionally included in this study to further evaluate the model’s generalization and convergence performance, which are often overlooked in prior research.
	Since ground-truth trajectories of Seqs. 11–21 are not publicly available, we generate pseudo ground-truth trajectories using ORB-SLAM3 (Stereo) \cite{campos2021orbslam3} for performance evaluation.
	To further validate the practical effectiveness of our method, we create the MIAS-Odom dataset, collected using a handheld setup, as illustrated in Fig.~\ref{fig.handheld_setup}. 
	This dataset consists of six (three indoor and three outdoor) video sequences with a total of 11,608 images captured under challenging conditions that are rarely covered by public datasets. These conditions include large rotational motions and frequent camera shakes, which typically lead to substantial rotational errors, as well as overexposure, poor illumination, and motion blur, all of which degrade the reliability of per-pixel photometric supervision.
	For ground-truth trajectory acquisition, we construct a system consisting of multiple hardware-synchronized sensors, including binocular cameras, a Livox MID-360 solid-state LiDAR, and an onboard IMU. Camera trajectories are obtained using a tightly coupled LiDAR-inertial odometry framework \cite{xu2022fast} that fuses LiDAR features with IMU measurements to achieve accurate and robust pose estimation.
	
	Depth estimation performance is evaluated on five public datasets: KITTI \cite{geiger2012we}, DDAD \cite{guizilini20203d}, nuScenes \cite{saxena2008make3d}, Make3D \cite{saxena2008make3d}, and DIML \cite{cho2021diml}. For the KITTI dataset, we adopt the Eigen split \cite{eigen2014depth}, which comprises 39,180 monocular triplets for training, 4,424 images for validation, and 697 images for testing. The DDAD dataset is split into a training set of 12,650 images and a test set of 3,950 images. For the nuScenes dataset, we use 79,760 image triplets for training, and evaluate the model’s performance on 6,019 front-camera images. As the Make3D and DIML datasets do not provide either stereo image pairs or monocular sequences for unsupervised model training, they are used exclusively to quantify the model's generalizability.

	\begin{table}[!t]
		\begin{center}
			\settablefont
			\caption{
				Quantitative visual odometry results on the test sets of the KITTI Odometry dataset \cite{geiger2013vision}. The best results within each method category are shown in bold type. `S' and `M' indicate training with stereo image pairs and monocular video sequences, respectively. Results for \cite{mur2017orb, campos2021orbslam3} are reported in monocular settings with loop closure enabled.
			}
			\label{tb.kitti_odom}
			
			\begin{tabular}{l@{\hspace{7pt}}c@{\hspace{7pt}}c@{\hspace{7pt}}|ccc|ccc}
				\toprule
				\multirow{2}{*}{Methods} & \multirow{2}{*}{Year} & \multirow{2}{*}{Data} & \multicolumn{3}{c|}{Seq. 09}  & \multicolumn{3}{c}{Seq. 10}   \\
				\cline{4-9}
				&&& $e_t$ & $e_r$ & ATE& $e_t$ & $e_r$ & ATE \\
				
				\hline
				\rowcolor{gray!20}\multicolumn{9}{c}{Geometric-Based Methods}\\
				DSO \cite{engel2017direct} & 2017 & --	& -- & -- & 52.22 & -- & -- & 11.09  \\
				ORB-SLAM2 \cite{mur2017orb} & 2017 & --	& 2.88 & \textbf{0.25} & 8.39 & 3.30 & \textbf{0.30} & 6.63  \\
				ORB-SLAM3 \cite{campos2021orbslam3} & 2021 & --	&	\textbf{2.35} & 0.28 & \textbf{6.06} & \textbf{2.24} & 0.31 & \textbf{4.97}  \\
				
				\hline
				\rowcolor{gray!20}\multicolumn{9}{c}{Hybrid Methods}\\
				DVSO \cite{yang2018deep} & 2018  & S   & 0.83 & \textbf{0.21} & -- & 0.74 & \textbf{0.21} & --  \\
				D3VO \cite{yang2020d3vo} & 2020  & S  & \textbf{0.78} & -- & -- & \textbf{0.62} & -- & -- \\
				TrianFlow \cite{zhao2020towards} & 2020 & M  & 6.93 & 0.44 & -- & 4.66 & 0.62 & -- \\
				pRGBD \cite{tiwari2020pseudo} & 2020 & M  & 4.20 & 1.00 & 11.97 & 4.40 & 1.60 & 6.35 \\
				DF-VO \cite{zhan2020visual} & 2021  & M  & 2.40 & 0.24 & 8.36 & 1.82 & 0.38 & 3.13 \\
				SC-Depth \cite{bian2021unsupervised} & 2021  & M  & 5.08 & 1.05 & 13.40 & 4.32 & 2.34 & 7.99 \\
				Sun \textit{et al.} \cite{sun2022improving} & 2022 & M  & -- & -- & 45.95 & -- & -- & 8.11 \\
				Wang \textit{et al.} \cite{wang2024self} & 2024  & M  & 1.86 & 0.29 & \textbf{6.06} & \textbf{1.55} & \textbf{0.31} & \textbf{2.64} \\
				
				\hline
				\rowcolor{gray!20}\multicolumn{9}{c}{Learning-Based Methods}\\
				Zhan \textit{et al.} \cite{zhan2018unsupervised} & 2018	&  S & 11.89 & 3.60 & 52.12 & 12.82 & 3.41 & 24.70 \\
				UndeepVO \cite{li2018undeepvo} & 2018 &  S & 7.01 & 3.60 & -- & 10.63 & 4.60 & -- \\
				SfMLearner \cite{zhou2017unsupervised} & 2017 &  M  & 19.15	& 6.82 & 77.79 & 40.40 & 17.69 & 67.34 \\
				Monodepth2 \cite{godard2019digging} & 2019	&  M 	& 17.17  & 3.85 & 76.22  & 11.68 &5.31  &20.35 \\
				LTMVO \cite{zou2020learning} & 2020 &  M 	& 3.49	& 1.00 & 11.30  & 5.81 &1.80  &11.80 \\
				SC-Depth \cite{bian2021unsupervised} & 2021  &  M 	& 7.31	&3.05 &23.56  & 7.79 &4.90  &12.00 \\
				MLF-VO \cite{jiang2022mlfvo} & 2022  &  M 	& 3.90 &1.41 &	9.86  & 4.88 & 1.38	& 7.36 \\
				MotionHint \cite{wang2022motionhint} & 2022  &  M 	& 8.18 &1.50 &17.82  & 8.10 & 2.74  &11.63 \\
				SCIPaD \cite{feng2024scipad} & 2024 &  M & 7.43	& 2.46 & 26.15 & 9.82 &3.87  &15.51 \\
				Lite-SVO \cite{wei2024lite} & 2024 &  M &7.20  &1.50 & 33.82 & 8.49 &2.40  &15.45 \\
				Manydepth2 \cite{zhou2025manydepth2} & 2025 &  M   &  7.01 &1.76 & -- & 7.29 & 2.65 & -- \\
				
				\rowcolor{orange!20}\textbf{DiMoDE (Ours)} & -- &  M  & \textbf{2.86}	&\textbf{0.74} & \textbf{9.83} & \textbf{4.20} &\textbf{1.22}  &\textbf{5.81} \\
				\bottomrule
				\toprule
				\multirow{2}{*}{Methods} & \multirow{2}{*}{Year} & \multirow{2}{*}{Data} & \multicolumn{3}{c|}{Seqs. 11-20}  & \multicolumn{3}{c}{Seq. 21}   \\
				\cline{4-9}
				&&& $e_t$ & $e_r$ & ATE& $e_t$ & $e_r$ & ATE \\
				\hline
				\rowcolor{gray!20}\multicolumn{9}{c}{Geometric-Based/Hybrid Methods}\\
				ORB-SLAM3 \cite{campos2021orbslam3} & 2021 & --	&	\textbf{4.05} & \textbf{0.38} & \textbf{13.46} & 94.16 & 2.38  & 829.27 \\
				TrianFlow \cite{zhao2020towards} & 2020 & M  & 18.77 & 5.57 &	76.96 & \textbf{17.56} & \textbf{1.99}	& \textbf{256.09} \\
				DF-VO \cite{zhan2020visual} & 2021  & M & 11.39 & 3.91 & 62.11 &	43.67&  13.84&  1491.25 \\
				\hline
				\rowcolor{gray!20}\multicolumn{9}{c}{Learning-Based Methods}\\
				Monodepth2 \cite{godard2019digging} & 2019	&  M 	& 16.16  & 5.64 & 53.26  & 16.88 &2.57  &526.75 \\
				SC-Depth \cite{bian2021unsupervised} & 2021  &  M 	& 15.52 & 5.43 & 55.31 & 44.79 & 4.50 & 800.90 \\
				SCIPaD \cite{feng2024scipad} & 2024 &  M & 7.98	& 3.65 & 35.78 & 16.27 &2.40  &\textbf{510.05} \\
				Manydepth2 \cite{zhou2025manydepth2} & 2025 &  M & 12.91 & 4.78 & 39.82 & 23.79  & 4.01	& 571.64  \\
				
				\rowcolor{orange!20}\textbf{DiMoDE (Ours)} & -- &  M  & 7.60	&\textbf{2.80} & \textbf{26.64} & 16.69 &\textbf{2.32}  &526.86 \\
				\bottomrule
			\end{tabular}
		\end{center}
	\end{table}
	
	\subsection{Evaluation Metrics}
	\label{Sect.metrics}
	Following the previous study \cite{sturm2012benchmark}, we adopt the average translational error $e_t$ (\%), average rotational error $e_r$ (${}^{\circ}/100$m), and the absolute trajectory error (ATE, in meters) to quantify pose estimation performance. Additionally, adhering to the study \cite{godard2019digging}, we quantify depth estimation performance using the mean absolute relative error (Abs Rel), mean squared relative error (Sq Rel), root mean squared error (RMSE), root mean squared log error (RMSE log), and the accuracy under specific thresholds ($\delta_i < 1.25^i$, where $i = 1, 2, 3$).
	
	\subsection{Monocular Visual Odometry Results}
	\label{Sect.vo_results}
	
	This subsection presents a comprehensive evaluation of monocular visual odometry performance. We compare DiMoDE with a wide range of SoTA approaches, including learning-based methods \cite{zhou2017unsupervised, zhan2018unsupervised, li2018undeepvo, godard2019digging, zou2020learning, bian2021unsupervised, jiang2022mlfvo, wang2022motionhint, feng2024scipad, wei2024lite}, geometry-based methods \cite{engel2017direct, mur2017orb, campos2021orbslam3}, and hybrid methods \cite{yang2018deep, yang2020d3vo, zhao2020towards, tiwari2020pseudo, zhan2020visual, bian2021unsupervised, sun2022improving, wang2024self}.
	
\begin{figure}[!t]
		\centering
		\includegraphics[width=0.49 \textwidth]{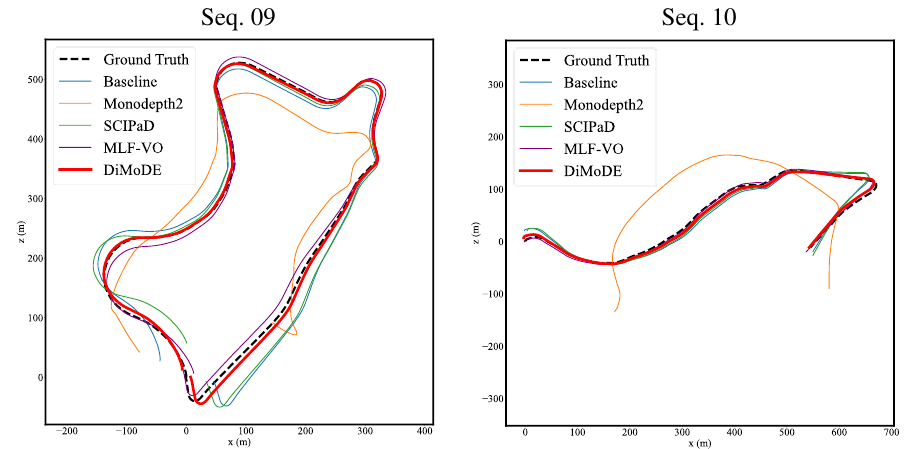}
		\caption{Comparisons of trajectories on the test sets of the KITTI Odometry dataset \cite{geiger2013vision}. All predicted trajectories are aligned with the ground truth using a 7-DoF similarity transformation.}
		\label{fig.traj3}
	\end{figure}

	Quantitative results on the KITTI Odometry dataset \cite{geiger2013vision} are presented in Table~\ref{tb.kitti_odom}. DiMoDE achieves SoTA performance among learning-based methods on the standard test set (Seqs. 09 and 10), exhibiting the lowest trajectory drift (see Fig.~\ref{fig.traj3}). However, evaluation on Seqs. 09-10 alone is insufficient to fully demonstrate the model's generalizability to unseen environments. To this end, we conduct additional comparative experiments on the remaining 11 sequences (Seqs. 11-21) in the dataset. Across these sequences, DiMoDE consistently achieves the highest average accuracy among all learning-based approaches. The results for Seq. 21 are reported separately due to its extreme difficulty, which causes all existing methods to fail. Furthermore, DiMoDE delivers competitive accuracy compared to geometry-based and hybrid approaches \cite{campos2021orbslam3, zhan2020visual}, significantly narrowing the performance gap between fully learning-based models and those relying on iterative optimization. 
	
	\begin{figure}[!t]
		\centering
		\includegraphics[width=0.49 \textwidth]{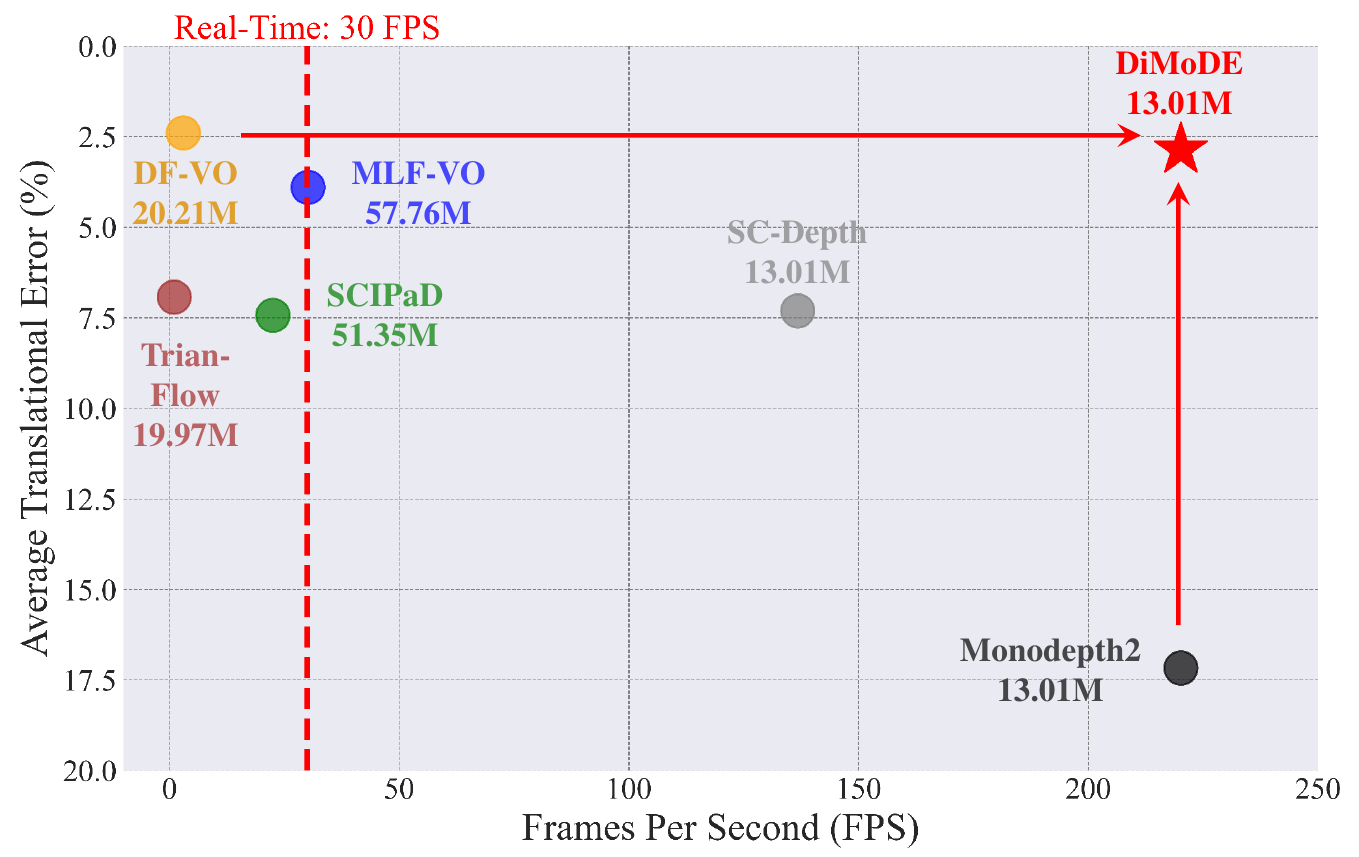}
		\caption{Comparison of storage overhead and inference speed among representative learning-based and hybrid visual odometry methods.}
		\label{fig.overhead}
	\end{figure}
	
	Moreover, we compare representative learning-based and hybrid methods in terms of storage overhead and inference speed, two crucial factors for practical deployment that have received relatively limited attention in previous studies. 
	As shown in Fig.~\ref{fig.overhead}, although hybrid methods \cite{zhao2020towards, zhan2020visual} generally outperform learning-based approaches due to the incorporation of back-end optimization, this advantage comes at the expense of significantly reduced inference speed, making them less suitable for real-world applications. 
	Previous learning-based approaches \cite{zou2020learning, jiang2022mlfvo, feng2024scipad} adopt increasingly sophisticated PoseNet architectures for improved accuracy. Nevertheless, such designs also incur substantial computational and memory costs, making real-time inference infeasible, even on high-end GPUs. In contrast, DiMoDE achieves competitive performance with SoTA hybrid methods while relying solely on a lightweight PoseNet architecture, thereby satisfying real-time inference requirements.
	
	\begin{table}[!t]
		\begin{center}
			\settablefont
			\caption{
				Quantitative visual odometry results on Seqs. 00, 02, 05, and 08 of the KITTI Odometry dataset \cite{geiger2013vision}. The best results within each method category are shown in bold type. `M' indicates training on monocular video sequences. Results for \cite{mur2017orb} are reported in monocular settings with loop closure enabled.
			}
			\label{tb.kitti_training_detail}
			
			\begin{tabular}{l@{\hspace{7pt}}c@{\hspace{7pt}}c@{\hspace{7pt}}|ccc|ccc}
				\toprule
				\multirow{2}{*}{Methods} & \multirow{2}{*}{Year} & \multirow{2}{*}{Data} & \multicolumn{3}{c|}{Seq. 00}  & \multicolumn{3}{c}{Seq. 02}   \\
				\cline{4-9}
				&&& $e_t$ & $e_r$ & ATE& $e_t$ & $e_r$ & ATE \\
				\hline
				\rowcolor{gray!20}\multicolumn{9}{c}{Geometric-Based/Hybrid Methods}\\
				ORB-SLAM3 \cite{campos2021orbslam3} & 2021 & --	&	3.13 & \textbf{0.46} & \textbf{8.70} & 5.10 & 0.55 & 25.11 \\
				DF-VO \cite{zhan2020visual}  & 2021 & M &\textbf{2.33} & 0.63 & 14.45 & \textbf{3.24} & \textbf{0.49} & \textbf{19.69} \\
				\hline
				\rowcolor{gray!20}\multicolumn{9}{c}{Learning-Based Methods}\\
				Monodepth2 \cite{godard2019digging} & 2019	&  M 	&11.40 & 3.26 & 107.51 & 6.42 & 1.51 & 83.63 \\
				SC-Depth \cite{bian2021unsupervised} & 2021  &  M 	&11.01 & 3.39 & 93.04 & 6.74 & 1.96 & 70.37 \\
				SCIPaD \cite{feng2024scipad} & 2024 &  M & 6.17 & 2.28 & 43.17 & 4.79 & 1.66 & 50.72 \\
				Manydepth2 \cite{zhou2025manydepth2} & 2025 &  M & 10.12 & 3.59 & 72.42 & 14.30 & 3.31	& 94.70 \\
				\rowcolor{orange!20}\textbf{DiMoDE (Ours)} & -- &  M  &\textbf{2.28} & \textbf{0.51} & \textbf{10.77} & \textbf{2.32} & \textbf{0.56} & \textbf{17.60}  \\
				\bottomrule
				\toprule
				\multirow{2}{*}{Methods} & \multirow{2}{*}{Year} & \multirow{2}{*}{Data} & \multicolumn{3}{c|}{Seq. 05}  & \multicolumn{3}{c}{Seq. 08}   \\
				\cline{4-9}
				&&& $e_t$ & $e_r$ & ATE& $e_t$ & $e_r$ & ATE \\
				\hline
				\rowcolor{gray!20}\multicolumn{9}{c}{Geometric-Based/Hybrid Methods}\\
				ORB-SLAM3 \cite{campos2021orbslam3} & 2021 & --	& 3.03 & 0.43 & 6.57 & 15.02 & \textbf{0.28} & 50.28 \\
				DF-VO \cite{zhan2020visual}  & 2021 & M & \textbf{1.09} & \textbf{0.25} & \textbf{3.61} & \textbf{2.18} & 0.32 & \textbf{7.63} \\
				\hline
				\rowcolor{gray!20}\multicolumn{9}{c}{Learning-Based Methods}\\
				Monodepth2 \cite{godard2019digging} & 2019	&  M 	& 7.21 & 3.11 & 42.35 & 7.08 & 2.44 & 56.35 \\
				SC-Depth \cite{bian2021unsupervised} & 2021  &  M 	& 6.70 & 2.38 & 40.56 & 8.11 & 2.61 & 56.15 \\
				SCIPaD \cite{feng2024scipad} & 2024 &  M & 3.77 & 1.75 & 13.51 & 4.58 & 1.83 & 22.03 \\
				Manydepth2 \cite{zhou2025manydepth2} & 2025 &  M & 6.06 & 2.52 & 29.04 & 8.56 & 3.02 & 61.90 \\
				\rowcolor{orange!20}\textbf{DiMoDE (Ours)} & -- &  M  & \textbf{1.63} & \textbf{0.62} & \textbf{6.47} & \textbf{2.10} & \textbf{0.65} & \textbf{10.01} \\
				
				\bottomrule
				
			\end{tabular}
		\end{center}
	\end{table}
	
	\begin{figure}[t!]
		{
			\setlength{\abovecaptionskip}{5pt}
			\centering
			\includegraphics[width=0.49 \textwidth]{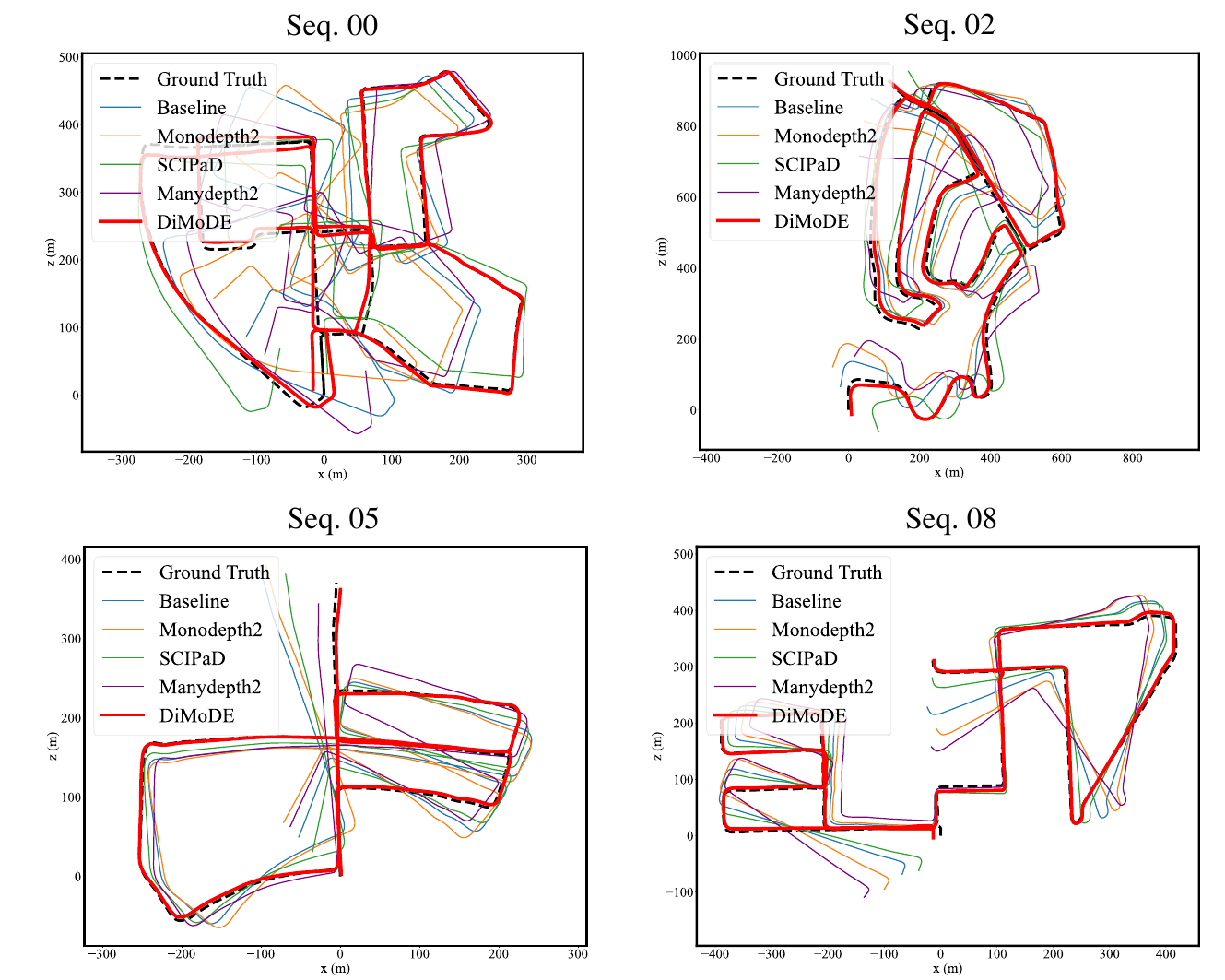}
			\caption{Comparisons of trajectories on several long-term sequences in the train set of the KITTI Odometry dataset \cite{geiger2013vision}. All predicted trajectories are aligned with the ground truth using a 7-DoF similarity transformation.}
			\label{fig.trajtraining}
		}
	\end{figure}
	
	To validate whether the DiMoDE framework provides more effective supervisory signals for pose estimation, we report the results on several long-term sequences (Seqs. 00, 02, 05, and 08) from the training set. As shown in Table~\ref{tb.kitti_training_detail}, previous methods, limited by the accuracy of their estimated ego-motion components, tend to accumulate significant errors over these long-term sequences, even when trained on them. In contrast, DiMoDE significantly alleviates this issue, outperforming the previous SoTA method, with ATE reductions of 75.05\%, 65.30\%, 52.11\%, and 54.56\%, respectively. Moreover, DiMoDE achieves performance competitive with approaches that rely on iterative optimization. Qualitative comparisons presented in Fig. \ref{fig.trajtraining} show that DiMoDE exhibits the least trajectory drift, highlighting its effectiveness in reducing error accumulation over complex and long-term trajectories.
	
	\begin{table}[!t]
		\begin{center}
			\settablefont
			\caption{
				Quantitative visual odometry results on Seqs. 11-20 of the KITTI Odometry dataset \cite{geiger2013vision}. The best results are shown in bold type. For fair comparisons, the DepthNet and FlowNet in DF-VO \cite{zhan2020visual}, as well as the PoseNet in all learning-based methods, are finetuned for the same number of iterations on these sequences.
			}
			\label{tb.kitti_add}
			
			\begin{tabular}{l|ccccc}
				\toprule
				\multirow{2}{*}{Methods}& $\text{Seq.11}$ & $\text{Seq. 12}$  & $\text{Seq. 13}$  & $\text{Seq. 14}$  & $\text{Seq. 15}$\\
				\cline{2-6}
				& ATE & ATE & ATE & ATE & ATE \\
				\hline
				
				DF-VO \cite{zhan2020visual}  &4.45 & 50.54 & 17.24 & 3.04 & 6.57 \\
				SC-Depth \cite{bian2021unsupervised}	&24.55 & 14.08 & 47.30 & 9.26 & 41.72 \\
				SCIPaD \cite{feng2024scipad}  &12.94 &	22.50 &	39.96 &	\textbf{1.91} &	20.46 \\
				Manydepth2 \cite{zhou2025manydepth2}	&10.04 &	20.30 &	73.38 &	5.56 &	11.96 \\
				\rowcolor{orange!20}\textbf{DiMoDE (Ours)} 
				&\textbf{4.26} &	\textbf{12.68} &	\textbf{8.23} &	2.17 &	\textbf{9.74}  \\
				\bottomrule
				\toprule
				\multirow{2}{*}{Methods}& $\text{Seq.16}$ & $\text{Seq. 17}$  & $\text{Seq. 18}$  &  \text{Seq. 19} & \text{Seq. 20}  \\
				\cline{2-6}
				& ATE & ATE & ATE & ATE & ATE \\
				\hline
				DF-VO \cite{zhan2020visual}  &\textbf{4.88} & 10.31 & 13.60 & 58.89 & 16.40 \\
				SC-Depth \cite{bian2021unsupervised}	&16.96 & 7.40 & 53.67 & 202.96 & 14.86 \\
				SCIPaD \cite{feng2024scipad} &14.99 &	4.97 &	36.65 &	150.58 & 11.92 \\
				Manydepth2 \cite{zhou2025manydepth2}	&28.51 & 6.92 & 15.11 & 124.18 & 15.83 \\
				\rowcolor{orange!20}\textbf{DiMoDE (Ours)} 
				&6.17 &	\textbf{1.55} &	\textbf{9.97} &	\textbf{25.23} &	\textbf{8.78} \\
				
				\bottomrule
				
			\end{tabular}
		\end{center}
	\end{table}

	\begin{figure}[t!]
		{
			\setlength{\abovecaptionskip}{5pt}
			\centering
			\includegraphics[width=0.49 \textwidth]{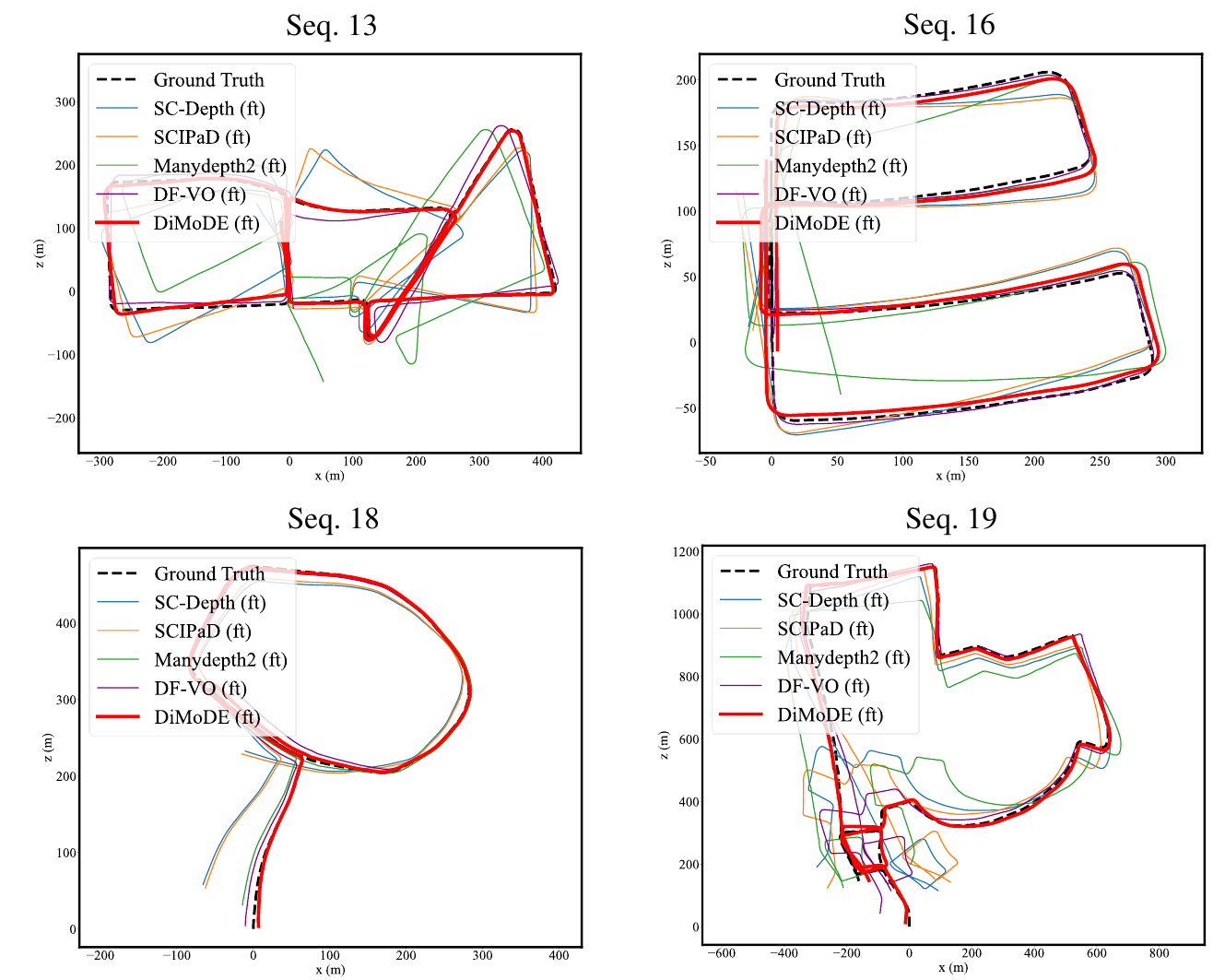}
			\caption{
				Comparisons of trajectories produced by models finetuned on the additional sequences of the KITTI Odometry dataset (`ft' denotes finetuned models). All predicted trajectories are aligned with the ground truth using a 7-DoF similarity transformation.
			}
			\label{fig.trajadd}
		}
	\end{figure}
	
\begin{figure}[t!]
		{
			\setlength{\abovecaptionskip}{5pt}
			\centering
			\includegraphics[width=0.49 \textwidth]{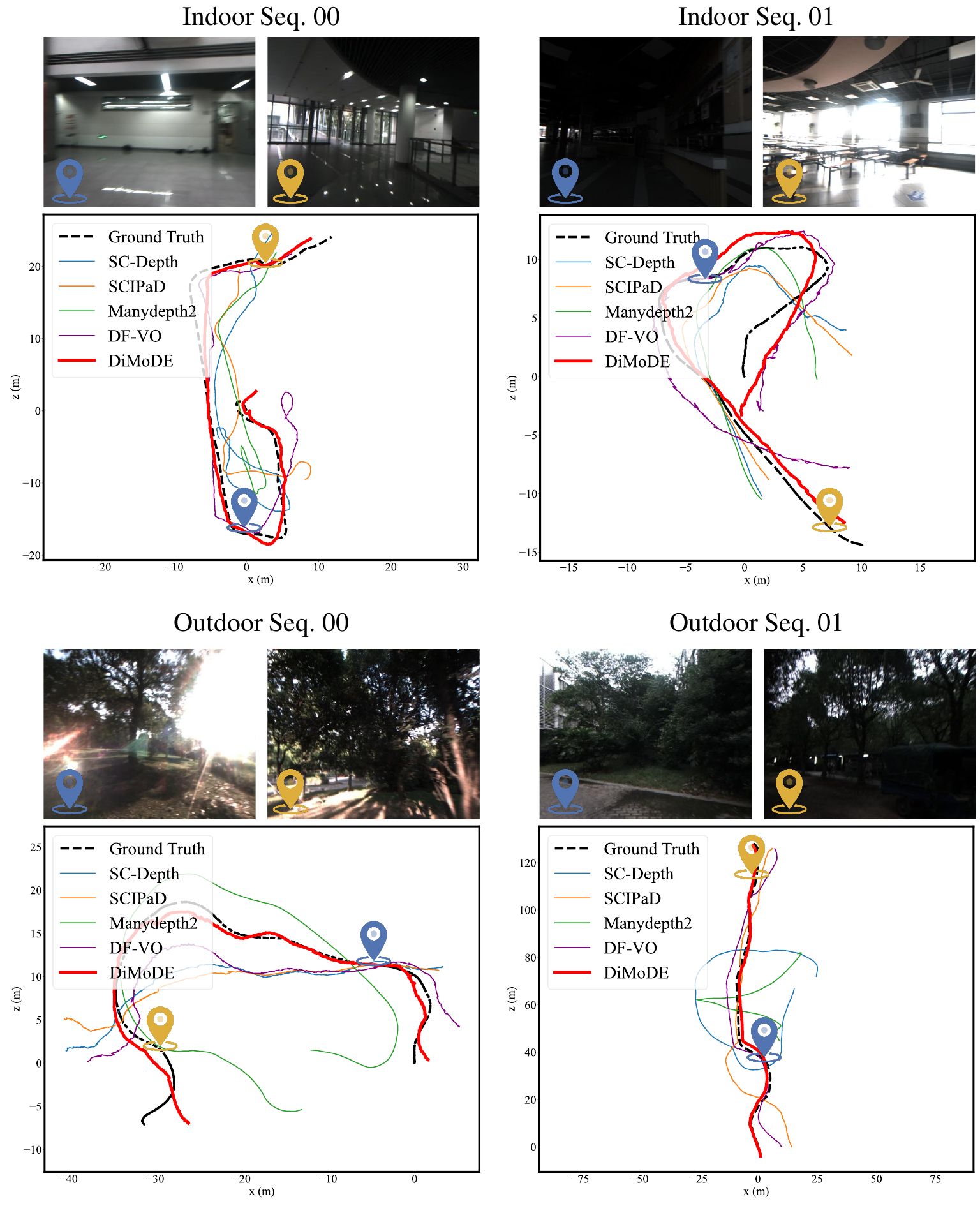}
			\caption{Comparisons of trajectories produced by models finetuned on the newly created MIAS-Odom dataset. All predicted trajectories are aligned with the ground truth using a 7-DoF similarity transformation.}
			\label{fig.trajcreated}
		}
	\end{figure}

	Furthermore, a key strength of unsupervised learning-based approaches lies in their inherent ability to adapt the model to previously unseen environments, which is crucial for real-world applications. To validate the effectiveness of the method in this regard, we conduct additional finetuning experiments on Seqs. 11–21 of the KITTI Odometry dataset and on our MIAS-Odom dataset, respectively, by training each model for five additional epochs. 
	As shown in Table \ref{tb.kitti_add} and Fig.~\ref{fig.trajadd}, while the other methods \cite{bian2021unsupervised, feng2024scipad, zhou2025manydepth2} continue to yield unsatisfactory results after finetuning on Seqs. 11–21 of the KITTI Odometry dataset, DiMoDE enables the estimated trajectories to align closely with the pseudo ground truth. This improvement corroborates our analyses in Sect.~\ref{sect.method1} and demonstrates that decomposing translational components effectively mitigates error propagation and improves convergence, thereby enabling more effective adaptation to previously unseen environments. 
	As shown in Table~\ref{tb.mias_odom} and Fig.~\ref{fig.trajcreated}, DiMoDE demonstrates significantly more robust ego-motion estimation on the MIAS-Odom dataset, effectively handling a variety of real-world challenges, such as overexposure, low illumination, severe motion blur, frequent camera shake, and large-degree turns. In contrast, prior methods consistently fail under these adverse conditions.
	However, our experimental results suggest that such satisfactory results can only be achieved when the challenging sequences are included in the training set. When a portion or all of these sequences are reserved exclusively for testing, all models, including DiMoDE, fail to generalize effectively. This generalization limitation stands in sharp contrast to the results observed on the KITTI Odometry dataset, and is further analyzed in Sect.~\ref{Sect.discussion}.
	
	\begin{table}[!t]
		\begin{center}
			\settablefont
			\caption{Quantitative visual odometry results on our newly created MIAS-Odom dataset. The best results are shown in bold type. For fair comparisons, the DepthNet and FlowNet in DF-VO \cite{zhan2020visual}, as well as the PoseNet in all learning-based methods, are finetuned for the same number of iterations on these sequences.}
			\label{tb.mias_odom}
			
			\begin{tabular}{l|ccc}
				\toprule
				\multirow{2}{*}{Methods}& $\text{Indoor Seq.00}$ & $\text{Indoor Seq. 01}$  & $\text{Indoor Seq. 02}$  \\
				& ATE & ATE & ATE \\
				\hline
				DF-VO \cite{zhan2020visual} & 3.53 & 3.61 & 5.94 \\
				SC-Depth \cite{bian2021unsupervised}	&9.02	&5.20   &10.67    \\
				SCIPaD \cite{feng2024scipad}  &5.83	  &5.42   &8.82    \\
				Manydepth2 \cite{zhou2025manydepth2}	& 7.67	& 4.51  &  11.21 \\
				\rowcolor{orange!20}\textbf{DiMoDE (Ours)} 
				&\textbf{1.48}	&\textbf{3.02}   &\textbf{5.33}  \\
				\bottomrule
				\toprule
				\multirow{2}{*}{Methods}& $\text{Outdoor Seq.00}$ & $\text{Outdoor Seq. 01}$  & $\text{Outdoor Seq. 02}$  \\
				& ATE & ATE & ATE \\
				\hline
				DF-VO \cite{zhan2020visual} & 4.43 & 4.35 & 11.07 \\
				SC-Depth \cite{bian2021unsupervised}	&6.74	&27.63   &11.23    \\
				SCIPaD \cite{feng2024scipad} &7.32	&8.02   &9.70    \\
				Manydepth2 \cite{zhou2025manydepth2}	&6.36	&32.92   & 24.31  \\
				\rowcolor{orange!20}\textbf{DiMoDE (Ours)} 
				&\textbf{2.71}	&\textbf{2.22}   &\textbf{6.48}  \\
				
				\bottomrule
				
			\end{tabular}
		\end{center}
	\end{table}

	\subsection{Monocular Depth Estimation Results}
	\label{Sect.depth_results}
	
	This subsection presents comprehensive experimental results to evaluate the performance of DiMoDE in monocular depth estimation. We compare our method with representative single-frame approaches \cite{he2022ra, zhang2023lite, sun2023dynamo, sun2023sc, shao2024monodiffusion, zhang2024dcpi}, multi-frame approaches \cite{watson2021temporal, bangunharcana2023dualrefine, zhou2025manydepth2}, and robust depth estimation approaches \cite{yan2025synthetic, sun2025depth}. All the compared methods provide publicly available implementations, allowing us to conduct additional training and evaluation on the DDAD \cite{guizilini20203d} and nuScenes \cite{caesar2020nuscenes} datasets.
	
	\begin{table*}[!t]
		\begin{center}
			\settablefont
			\renewcommand{\arraystretch}{1}
			\caption{
				Quantitative comparison with SoTA methods on the KITTI \cite{geiger2012we} dataset. The best results for single-frame and multi-frame approaches are shown in bold type, respectively. The symbols $\uparrow$ and $\downarrow$ indicate that higher and lower values correspond to better performance, respectively.
			}
			\label{tb.depth_kitti}
			{
				\begin{tabular}{lccc|cccc|ccc}
					\toprule
					Method & Year &  Test frames & Resolution (pixels) & Abs Rel $\downarrow$ & Sq Rel $\downarrow$ & RMSE $\downarrow$ & RMSE log $\downarrow$ & $\delta <1.25$ $\uparrow$ & $\delta <1.25^2$ $\uparrow$ & $\delta <1.25^3$ $\uparrow$  \\
					\hline
					\rowcolor{gray!20}\multicolumn{11}{c}{Multi-Frame Methods}\\
					Manydepth \cite{watson2021temporal} & 2021 & 2(-1, 0)  & 640 $\times$ 192 &   0.098   & 0.770   &  4.459  &  0.176  &  0.900  &  0.965 &  0.983  \\
					DualRefine \cite{bangunharcana2023dualrefine} & 2023  & 2(-1, 0) & 640 $\times$ 192 &   0.105    &  0.787 & 4.544   &  0.183  & 0.891   &  0.964  &  0.983 \\
					Manydepth2 \cite{zhou2025manydepth2} & 2025 & 2(-1, 0)  & 640 $\times$ 192 &   \textbf{0.091}  &  \textbf{0.649}  & \textbf{4.232}  &  \textbf{0.170} &  \textbf{0.909} &  \textbf{0.968}  &  \textbf{0.984} \\
					\hline
					\rowcolor{gray!20}\multicolumn{11}{c}{Single-Frame Methods}\\
					D-HRNet \cite{he2022ra} & 2022 & 1  & 640 $\times$ 192  &  0.102 & 0.760 & 4.479 & 0.179 & 0.897 & 0.965 & 0.983 \\
					Lite-Mono \cite{zhang2023lite} & 2023 & 1  & 640 $\times$ 192    & 0.101 & 0.729 & 4.454 & 0.178 & 0.897 & 0.965 & 0.983 \\
					Dynamo-Depth \cite{sun2023dynamo} & 2023 & 1  & 640 $\times$ 192    & 0.112 & 0.758 & 4.505 & 0.183 & 0.873 & 0.959 & 0.984 \\
					SC-DepthV3 \cite{sun2023sc} & 2024 & 1  & 640 $\times$ 192    & 0.118 & 0.756 & 4.709 & 0.188 & 0.864 & 0.960 & \textbf{0.985} \\
					MonoDiffusion \cite{shao2024monodiffusion} & 2024 & 1  & 640 $\times$ 192 & 0.099 & 0.702 & 4.385 & 0.176 & 0.899 & 0.966 & 0.984 \\
					DCPI-Depth \cite{zhang2024dcpi} & 2025 & 1  & 640 $\times$ 192 & 0.097 &	0.666 &	4.388 &	0.173 &	0.898 &	0.966 &	0.985 \\
					\rowcolor{orange!20} \textbf{DiMoDE (Ours)} & - & 1  & 640 $\times$ 192   &\textbf{0.097} &\textbf{0.652} &\textbf{4.337} &\textbf{0.172} &\textbf{0.899} &\textbf{0.967} &\textbf{0.985} \\
					
					\bottomrule
				\end{tabular}
			}
		\end{center}
	\end{table*}
	
	\begin{table*}[!t]
		\begin{center}
			\settablefont
			\renewcommand{\arraystretch}{1}
			\caption{
				Quantitative comparison with SoTA methods on the DDAD \cite{guizilini20203d} dataset. The best results for single-frame and multi-frame approaches are shown in bold type, respectively. 
				The symbols $\uparrow$ and $\downarrow$ indicate that higher and lower values correspond to better performance, respectively.
			}
			\label{tb.depth_ddad}  
			{
				\begin{tabular}{lccc|cccc|ccc}
					\toprule
					Method & Year &  Test frames & Resolution (pixels) & Abs Rel $\downarrow$ & Sq Rel $\downarrow$ & RMSE $\downarrow$ & RMSE log $\downarrow$ & $\delta <1.25$ $\uparrow$ & $\delta <1.25^2$ $\uparrow$ & $\delta <1.25^3$ $\uparrow$  \\
					\hline
					\rowcolor{gray!20}\multicolumn{11}{c}{Multi-Frame Methods}\\
					Manydepth \cite{watson2021temporal} & 2021 & 2(-1, 0)  & 640 $\times$ 384 &   \textbf{0.178}  &   \textbf{3.455}  &  \textbf{15.409}  &   \textbf{0.268}  &   \textbf{0.753}  &   \textbf{0.908}  &   \textbf{0.961}  \\
					DualRefine \cite{bangunharcana2023dualrefine} & 2023 & 2(-1, 0) & 640 $\times$ 384  &   0.186  &   4.141  &  15.741  &   0.271  &   0.741  &   0.907  &   0.960  \\
					Manydepth2 \cite{zhou2025manydepth2} & 2025 & 2(-1, 0)  & 640 $\times$ 384 &   0.183  &   4.007  &  16.014  &   0.283  &   0.761  &   0.903  &   0.950  \\
					\hline
					\rowcolor{gray!20}\multicolumn{11}{c}{Single-Frame Methods}\\
					D-HRNet \cite{he2022ra} & 2022 & 1  & 640 $\times$ 384  &   0.198  &  8.124  &  16.316  &   0.289  &   0.798  &   0.914  &   0.956  \\
					Lite-Mono \cite{zhang2023lite}             & 2023  & 1  & 640 $\times$ 384   &	  0.175  &   6.425  &  16.687  &   0.272  &   0.799  &   0.920  &   0.961  \\
					Dynamo-Depth \cite{sun2023dynamo} & 2023  & 1 & 640 $\times$ 384 &   0.150  &   3.219  &  14.852  &   0.246  &   0.798  &   0.927  &   0.969  \\
					SC-DepthV3 \cite{sun2023sc} & 2024 & 1  & 640 $\times$ 384 & 0.142 & 3.031 & 15.868 & 0.248 & 0.813 & 0.922 & 0.963 \\
					MonoDiffusion \cite{shao2024monodiffusion} & 2024 & 1  & 640 $\times$ 384 &   0.163  &   6.362  &  15.922  &   0.255  &   0.815  &   0.928  &   0.966  \\
					DCPI-Depth \cite{zhang2024dcpi} & 2025 & 1  & 640 $\times$ 384 & 0.141 &	2.711 &	14.757 &	0.236 &	0.813 &	0.931 &	0.971 \\
					\rowcolor{orange!20} \textbf{DiMoDE (Ours)}  & - 	& 1	& 640 $\times$ 384    &   \textbf{0.134}  &   \textbf{2.685}  &  \textbf{14.119}  &   \textbf{0.230}  &   \textbf{0.831}  &   \textbf{0.934}  &   \textbf{0.972}  \\
					
					\bottomrule
				\end{tabular}
			}
		\end{center}
	\end{table*}
	
	\begin{table*}[!t]
		\begin{center}
			\settablefont
			\renewcommand{\arraystretch}{1}
			\caption{
				Quantitative comparison with SoTA methods on the nuScenes \cite{caesar2020nuscenes} dataset. The best results for single-frame, multi-frame, and robust depth estimation approaches are shown in bold type, respectively. The symbols $\uparrow$ and $\downarrow$ indicate that higher and lower values correspond to better performance, respectively. 
				Results for \cite{sun2025depth} are reported using optimal scale and shift to align predictions with the ground truth, rather than the median alignment used in other methods, which is necessary because the model predicts scale-shift-invariant depth and requires bias correction for alignment \cite{wang2025jasmine}.
			}
			\label{tb.depth_nuscenes}  
			{
				\begin{tabular}{lccc|cccc|ccc}
					\toprule
					Method & Year &  Test frames & Resolution (pixels) & Abs Rel $\downarrow$ & Sq Rel $\downarrow$ & RMSE $\downarrow$ & RMSE log $\downarrow$ & $\delta <1.25$ $\uparrow$ & $\delta <1.25^2$ $\uparrow$ & $\delta <1.25^3$ $\uparrow$  \\
					\hline
					\rowcolor{gray!20}\multicolumn{11}{c}{Robust Depth Estimation Methods}\\
					Syn2Real-Depth \cite{yan2025synthetic} & 2025 & 2(-1, 0) & 576 $\times$ 320 &   \textbf{0.152}  &   \textbf{1.884}  &   7.442  &   \textbf{0.244}  &   \textbf{0.813}  &   \textbf{0.930}  &   \textbf{0.968}  \\
					DepthAnything-AC \cite{sun2025depth} & 2025 & 1 & 512 $\times$ 288 &  0.208 &	3.230 &	\textbf{7.035} &	0.281 &	0.754 &	0.903 &	0.954
					\\
					\hline
					\rowcolor{gray!20}\multicolumn{11}{c}{Multi-Frame Methods}\\
					Manydepth \cite{watson2021temporal} & 2021 & 2(-1, 0)  & 512 $\times$ 288 &   0.264  &   3.418  &   9.316  &   0.351  &   0.621  &   0.836  &   0.924  \\
					DualRefine \cite{bangunharcana2023dualrefine} & 2023  & 2(-1, 0) & 512 $\times$ 288 
					&   \textbf{0.255}  &   \textbf{3.204}  &   \textbf{9.129}  &   \textbf{0.334}  &   \textbf{0.635}  &   \textbf{0.844}  &   \textbf{0.929}  \\
					Manydepth2 \cite{zhou2025manydepth2} & 2025 & 2(-1, 0)  & 512 $\times$ 288 &   0.287  &   4.964  &  11.132  &   0.380  &   0.612  &   0.812  &   0.902  \\
					\hline
					\rowcolor{gray!20}\multicolumn{11}{c}{Single-Frame Methods}\\
					D-HRNet \cite{he2022ra} & 2022 & 1  & 512 $\times$ 288 &  0.396 &	15.087 &	10.414 &	0.386 &	0.715 &	0.838 &	0.892 \\
					Lite-Mono \cite{zhang2023lite} & 2023 & 1  & 512 $\times$ 288 & 0.419 & 15.578 & 9.807 & 0.449 & 0.720 & 0.831 & 0.879 \\
					Dynamo-Depth \cite{sun2023dynamo} & 2023 & 1  & 512 $\times$ 288 & 0.179 & 2.118 & 7.050 & 0.271 & 0.787 & 0.896 & 0.940 \\
					SC-DepthV3 \cite{sun2023sc} & 2024 & 1  & 512 $\times$ 288 &   0.167  &   2.003  &   7.701  &   0.251  &   0.784  &   0.917  &   0.965  \\
					MonoDiffusion \cite{shao2024monodiffusion} & 2024 & 1  & 512 $\times$ 288 &   0.223  &   3.564  &   8.906  &   0.299  &   0.723  &   0.887  &   0.947  \\
					DCPI-Depth \cite{zhang2024dcpi} & 2025 & 1  & 512 $\times$ 288 & 0.157 &	1.795 &	7.192 &	0.255 &	0.790 &	0.914 &	0.959 \\
					\rowcolor{orange!20} \textbf{DiMoDE (Ours)} & -  & 1 & 512 $\times$ 288  &   \textbf{0.139}  &   \textbf{1.472}  &   \textbf{6.254}  &   \textbf{0.226}  &   \textbf{0.836}  &   \textbf{0.939}  &   \textbf{0.973}  \\
					\bottomrule
				\end{tabular}
			}
		\end{center}
	\end{table*}
	
	As shown in Table~\ref{tb.depth_kitti}, our method achieves competitive performance but does not surpass existing SoTA approaches on the KITTI dataset. This is primarily because our method is designed to impose additional constraints for robust joint learning under adverse conditions and to address the inconsistent optimization issue discussed in \eqref{eq.grad_z}. However, the KITTI dataset is collected under clear daytime conditions, with few close-range objects near the principal point. As such, it lacks the types of challenges that DiMoDE is specifically designed to handle, thereby limiting the extent to which its advantages can be fully demonstrated.
	
\begin{figure}[!t]
		{
			\setlength{\abovecaptionskip}{5pt}
			\centering
			\includegraphics[width=0.49\textwidth]{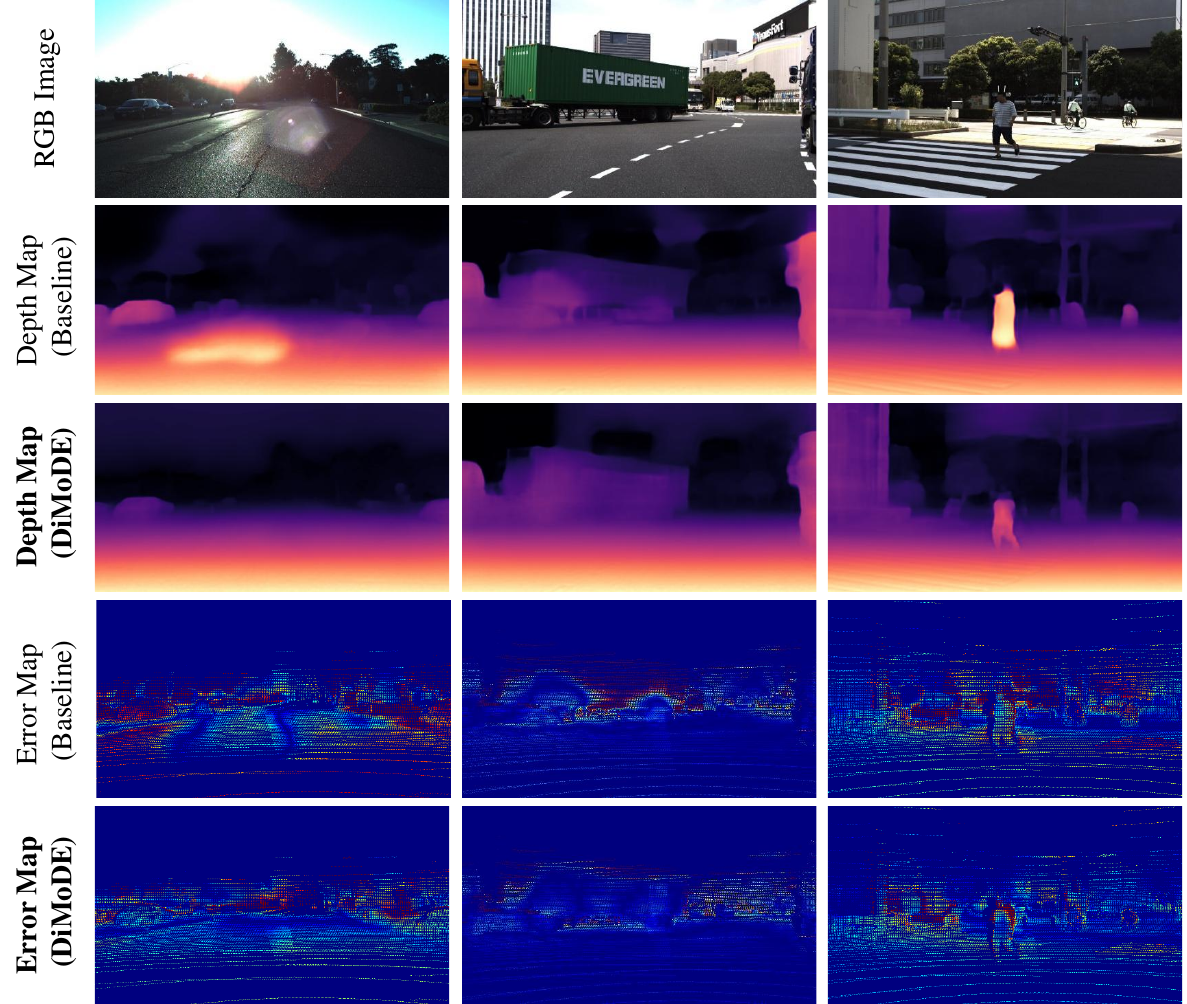}
			\caption{Qualitative results on the DDAD \cite{guizilini20203d} dataset.}
			\label{fig.depth_compare_ddad}
		}
	\end{figure}

\begin{figure}[!t]
		{
			\setlength{\abovecaptionskip}{5pt}
			\centering
			\includegraphics[width=0.49\textwidth]{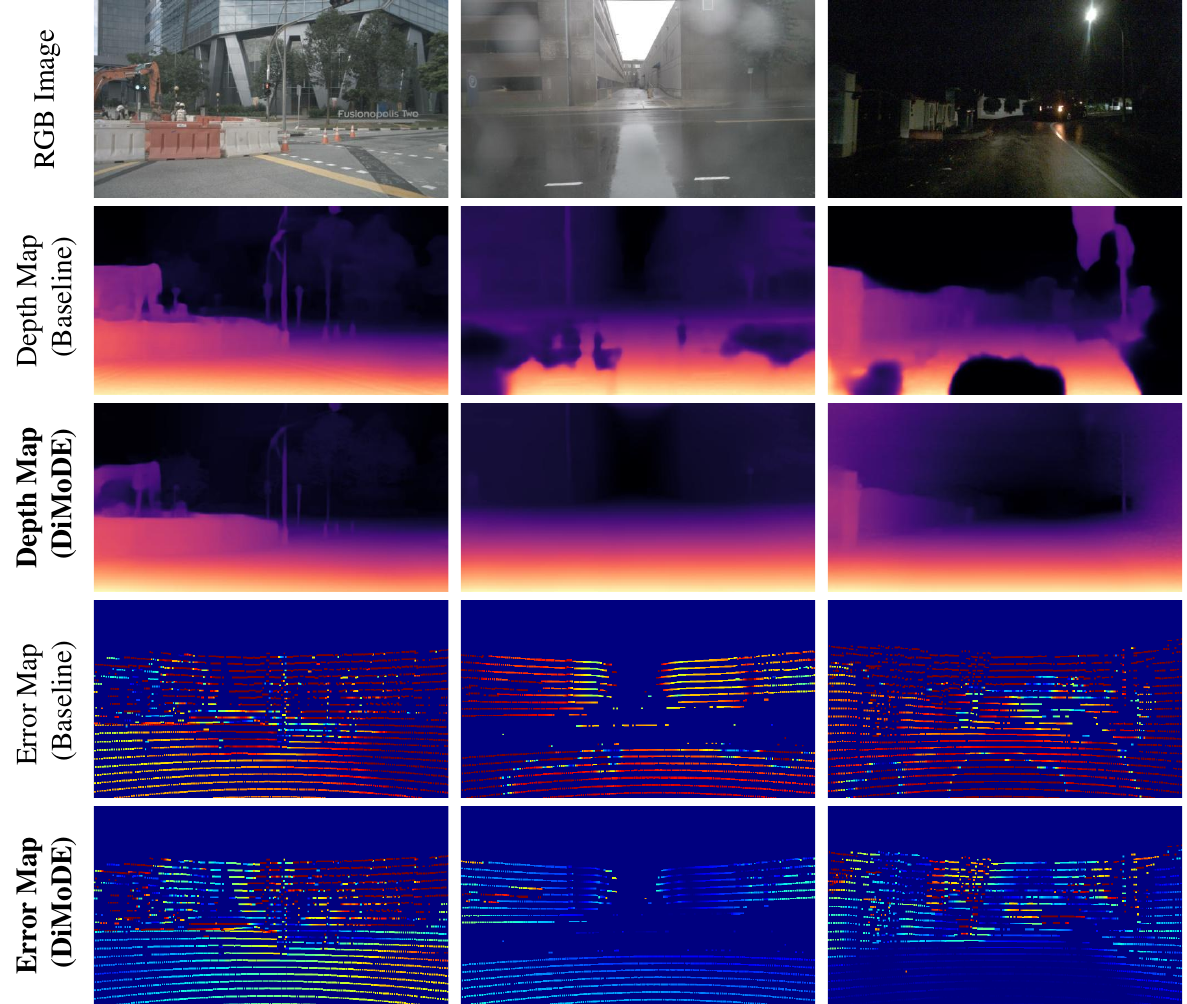}
			\caption{Qualitative results on the nuScenes \cite{caesar2020nuscenes} dataset.}
			\label{fig.depth_compare_nuscenes}
		}
	\end{figure}

	To further validate the effectiveness of the proposed method, we conduct experiments on the DDAD and nuScenes datasets, which contain more challenging scenarios, such as varying illumination, adverse weather, and structurally complex environments.
	As shown in Tables~\ref{tb.depth_ddad} and~\ref{tb.depth_nuscenes}, DiMoDE significantly outperforms previous leading methods \cite{sun2023dynamo, sun2023sc, zhang2024dcpi}. By comparison, several sophisticated networks \cite{he2022ra, zhang2023lite, bangunharcana2023dualrefine, shao2024monodiffusion, zhou2025manydepth2}, despite their strong performance on the KITTI dataset, exhibit reduced accuracy or even fail to converge on these more challenging datasets. This performance drop stems from their exclusive reliance on pixel-level supervisory signals, which are less robust under challenging conditions and often lead to suboptimal DepthNet predictions. In contrast, DiMoDE establishes a constraint cycle via \eqref{eq.lt} and \eqref{eq.lr}, which incorporates global geometric constraints into the training of DepthNet, enabling DiMoDE to effectively overcome the limitations of prior methods. 
	Qualitative results are shown in Figs.~\ref{fig.depth_compare_ddad} and \ref{fig.depth_compare_nuscenes}, demonstrating that DiMoDE consistently outperforms the baseline model across all aforementioned challenging conditions, which further supports our findings.
	Furthermore, we compare DiMoDE with two recent SoTA robust depth estimation methods \cite{yan2025synthetic} and \cite{sun2025depth} on the nuScenes dataset. Both approaches rely on specialized domain adaptation strategies, with the method \cite{sun2025depth} additionally incorporating a depth foundation model \cite{yang2025depth}. Remarkably, DiMoDE consistently delivers superior performance, underscoring its effectiveness and adaptability as a general-purpose framework for depth and ego-motion joint learning.

	
	Notably, we conduct additional evaluations on the DDAD dataset using models trained on either the KITTI or nuScenes dataset to investigate how variations in environmental conditions within the training data affect the model's generalizability. As shown in Table~\ref{table:cross_test}, methods that perform well on the KITTI dataset fail to generalize effectively to the DDAD dataset. In contrast, models converged on the nuScenes dataset demonstrate significantly better generalizability, achieving performance even comparable to those trained directly on the DDAD dataset.
	We attribute this phenomenon to the greater environmental complexity and diversity in the nuScenes dataset, which compels models to learn more robust and generalizable representations, rather than overfitting to repetitive, structured scenes in the KITTI dataset.
	This finding highlights the importance of developing generalizable frameworks that support robust learning across diverse environments, a central objective of DiMoDE.
	
	\begin{table}[!t]
		\begin{center}
			\settablefont
			\caption{Zero-shot depth estimation results on the DDAD \cite{saxena2008make3d} dataset. All models are trained on the KITTI or nuScenes dataset.}
			\label{table:cross_test}
			\begin{tabular}{lc|ccc}
				\toprule
				Method  &  Training Data  & Abs Rel $\downarrow$ & Sq Rel $\downarrow$ & $\delta <1.25$ $\uparrow$ \\ \hline
				\multirow{2}{*}{\makecell[c]{Dynamo-Depth \cite{sun2023dynamo}}} &  KITTI	&   0.191  &   4.091  &   0.696    \\
				&  nuScenes	&   0.153  &   3.418  &   0.785   \\
				\hline
				\multirow{2}{*}{\makecell[c]{SC-DepthV3 \cite{sun2023sc}}} &  KITTI	&   0.258  &   5.904  &   0.554  \\
				&  nuScenes	&   0.218  &   3.868   &   0.627    \\
				\hline
				\multirow{2}{*}{\makecell[c]{MonoDiffusion \cite{shao2024monodiffusion}}} &  KITTI	&   0.176  &   3.791  &   0.738  \\
				&  nuScenes	& 0.634  &  15.197  &   0.232    \\
				\hline
				\multirow{2}{*}{\makecell[c]{Syn2Real-Depth \cite{yan2025synthetic}}} &  KITTI	&   --  &   --  &   --  \\
				&  nuScenes	&   0.152  &   3.289  &   0.782  \\
				\hline
				\multirow{2}{*}{\makecell[c]{\textbf{DiMoDE (Lite-Mono)}}} &  KITTI	&   0.186  &   4.113  &   0.710  \\
				&  nuScenes	& \textbf{0.142}  &   \textbf{3.159}  &   \textbf{0.797} \\
				
				\bottomrule
			\end{tabular}
		\end{center}
	\end{table}
	
	\begin{table}[!t]
		\begin{center}
			\settablefont
			\caption{Zero-shot depth estimation results on the Make3D \cite{saxena2008make3d} and DIML \cite{cho2021diml} datasets. All models are trained on the KITTI dataset.}
			\label{table:make3d}
			\begin{tabular}{ll|cccc}
				\toprule
				Dataset&Method   & Abs Rel $\downarrow$ & Sq Rel $\downarrow$ & RMSE $\downarrow$  & RMSE log $\downarrow$ \\ \hline
				\multirow{6}{*}{\makecell[c]{\textbf{Make3D}}}
				& Lite-Mono \cite{zhang2023lite}  		& 0.305   & 3.060  & 6.981 & 0.158    \\
				& Dynamo-Depth \cite{sun2023dynamo}  		& 0.314  & 3.259  & 7.040  & 0.157   \\
				& SC-DepthV3 \cite{sun2023sc}   &   0.373 &   5.584 &   8.469 &   0.177  \\
				& MonoDiffusion \cite{shao2024monodiffusion}   		&   0.297  &  2.871   &   6.877 &   0.156   \\
				& DCPI-Depth \cite{zhang2024dcpi} &  	0.291    	&  2.944   &  6.817  &  \textbf{0.150}	   \\
				\rowcolor{orange!20}& \textbf{DiMoDE (Ours)}		 &  	\textbf{0.289}    	&  \textbf{2.692}   &  \textbf{6.634}  &  0.151	   \\
				\hline
				\multirow{6}{*}{\makecell[c]{\textbf{DIML}}} 
				& Lite-Mono \cite{zhang2023lite} & 0.173 & 0.271 & 1.108 & 0.239 \\
				& Dynamo-Depth \cite{sun2023dynamo} &   0.170  &   0.262  &   1.063  &   0.227 \\
				& SC-DepthV3 \cite{sun2023sc} &   0.213  &   0.407  &   1.274  &   0.275  \\
				& MonoDiffusion \cite{shao2024monodiffusion} & 0.166 & 0.256 & 1.084 & 0.232 \\
				& DCPI-Depth \cite{zhang2024dcpi} & 0.163 & 0.237 & 1.038 & 0.226 \\
				
				\rowcolor{orange!20}& \textbf{DiMoDE (Ours)}		 &   \textbf{0.160}  &   \textbf{0.237}  &   \textbf{1.033}  &   \textbf{0.224}  \\
				
				\bottomrule
			\end{tabular}
		\end{center}
	\end{table}
	
	\begin{figure}[!t]
		\centering
		\includegraphics[width=0.49\textwidth]{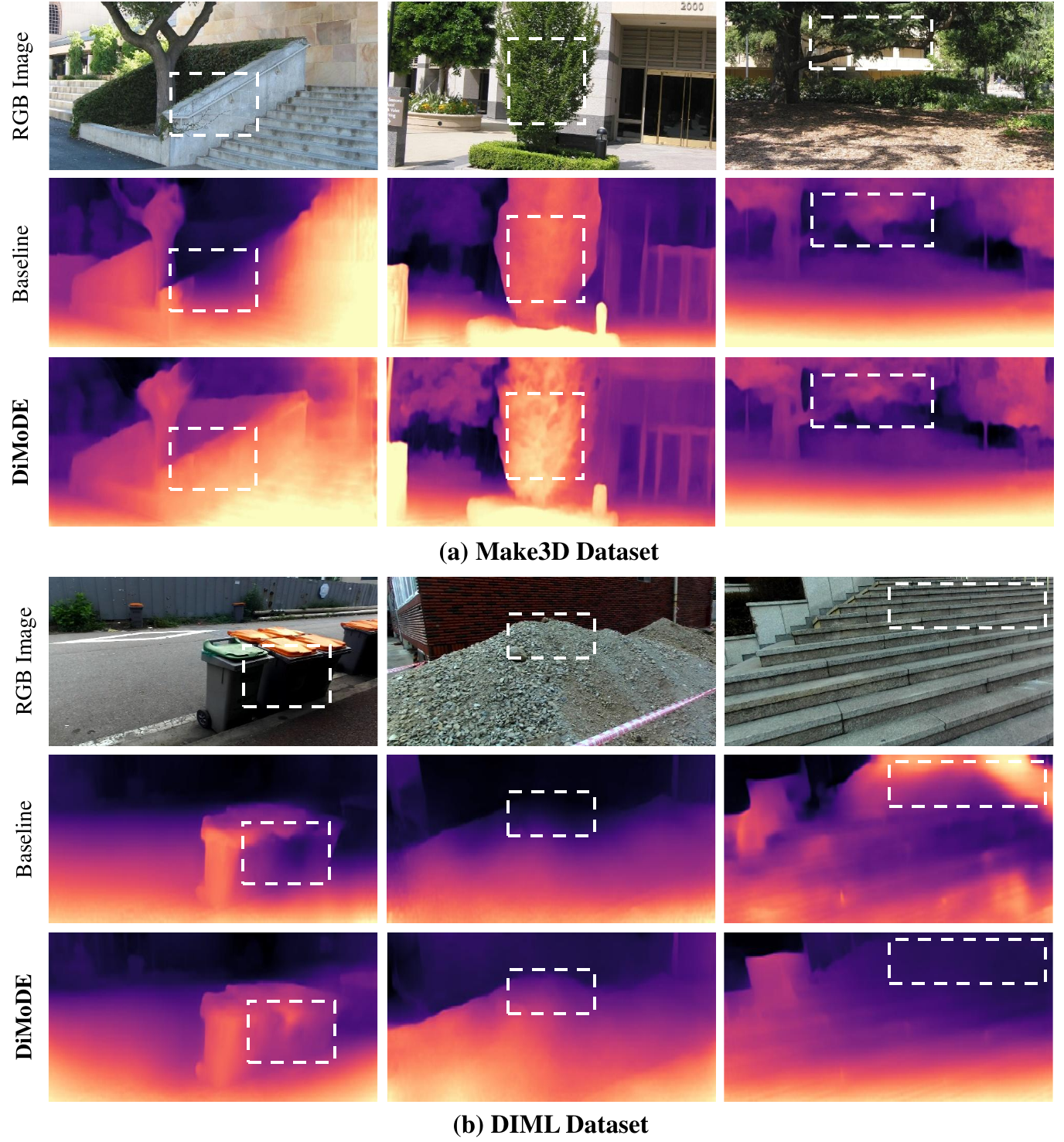}
		\caption{Qualitative zero-shot monocular depth estimation results on the Make3D \cite{saxena2008make3d} and DIML \cite{cho2021diml} datasets.}
		\label{fig.diml}
	\end{figure}
	
	Finally, we conduct zero-shot evaluations on the Make3D \cite{saxena2008make3d} and DIML \cite{cho2021diml} datasets using models trained on the KITTI dataset to evaluate our model’s generalizability to broader, more diverse scenarios. As shown in Table~\ref{table:make3d}, our method achieves slightly better quantitative metrics. Nonetheless, the qualitative comparisons presented in Fig.~\ref{fig.diml} reveal more pronounced improvements. In particular, the first row of Fig.~\ref{fig.diml}(a), as well as the second and third rows in Fig.~\ref{fig.diml}(b), demonstrate that our method effectively avoids overly distant predictions near the principal point, a common bias easily learned from the high-frequency scenes in the KITTI dataset.
	This finding further underscores the benefit of addressing perspective scaling effects, which helps maintain consistent optimization across the entire depth map. Moreover, additional qualitative examples in Fig.~\ref{fig.diml} show that our DepthNet produces fine-grained predictions in texture-rich regions and demonstrates greater reliability in texture-less areas, compared to the baseline model. These results collectively emphasize that the introduced geometric constraints provide complementary visual cues on top of the photometric supervision, thereby enhancing the robustness and fidelity of depth estimation.

	\subsection{Ablation Studies and Hyperparameter Selection}
	\label{sect.abl}
	
	The first ablation study, presented in Table~\ref{tb.vo_abl}, validates the effectiveness of the proposed constraints for refining PoseNet outputs. When both $\mathcal{L}_\text{axi}$ and $\mathcal{L}_\text{pla}$ are removed, a substantial degradation in visual odometry performance is observed, demonstrating the efficacy of imposing optical axis and imaging plane alignments. Furthermore, removing either $\mathcal{L}_\text{axi}$ or $\mathcal{L}_\text{pla}$ individually results in notable declines in both translation and rotation accuracy, comparable to those observed when both constraints are removed. These findings indicate that the two constraints play complementary roles in constraining rotation estimation, which are consistent with our analyses in Sect.~\ref{sect.method1}. Notably, effective refinement of translational components is only achieved when rotation is well constrained, thereby suppressing irregular rotational flows, as illustrated in Fig.~\ref{fig.residual}. Additionally, the conventional ResNet-based PoseNet is replaced with the architecture adopted in the study \cite{feng2024scipad} for both the full implementation and the baseline configuration. The results demonstrate that our framework is compatible with this alternative network and consistently yields significant performance gains.
	
	\begin{table}[!t]
		\begin{center}
			\settablefont
			\caption{Ablation study on the effectiveness of the additionally incorporated geometric constraints for refining PoseNet outputs on the KITTI Odometry dataset across different PoseNet architectures.}
			\label{tb.vo_abl}
			\begin{tabular}{lc|ccc|ccc}
				\toprule
				\multicolumn{2}{l|}{\multirow{2}{*}{Configuration}} 	& \multicolumn{3}{c|}{Seq. 09} & \multicolumn{3}{c}{Seq. 10}  \\           \cline{3-8} 
				&& $e_t$ & $e_r$ & ATE& $e_t$ & $e_r$  & ATE  \\ \hline 
				\multicolumn{2}{l|}{Full Implem. (ResNet-18)} &\textbf{2.86}	&\textbf{0.74} & \textbf{9.83} & \textbf{4.20} &\textbf{1.22}  &\textbf{5.81} \\ 
				\multicolumn{2}{l|}{Full Implem. (PoseNet from \cite{feng2024scipad})} &5.43 & 2.28 & 16.29 & 7.88 & 2.39 & 10.20  \\
				\hline
				\multicolumn{2}{l|}{w/o $\mathcal{L}_\text{axi}$ and $\mathcal{L}_\text{pla}$ (ResNet-18)} &5.84	&1.24	&26.39	&4.89	&1.49	&6.40 \\ 
				\multicolumn{2}{l|}{w/o $\mathcal{L}_\text{axi}$ (ResNet-18)} &5.29	&1.12	&23.63 &4.71	&1.31	&6.02 \\
				\multicolumn{2}{l|}{w/o $\mathcal{L}_\text{pla}$ (ResNet-18)} &4.86	&1.23	&21.96	&4.63	&1.59	&6.24 \\ 
				
				\hline
				\multicolumn{2}{l|}{Baseline (ResNet-18)}			&7.72	&1.71 & 35.71 & 8.12 &2.96  &11.65 \\
				\multicolumn{2}{l|}{Baseline (PoseNet from \cite{feng2024scipad})}			& 7.33	& 2.61 & 30.43 & 10.55 &3.89  &16.20 \\
				\bottomrule
				
			\end{tabular}
		\end{center}
	\end{table}
	
	\begin{table}[!t]
		\begin{center}
			\settablefont
			\caption{
				Ablation study on the efficacy of the geometric constraints from decomposed flows across different DepthNet architectures, compared with the cross-task consistency loss \cite{zou2018df} and the contextual-geometric depth consistency loss \cite{zhang2024dcpi}, both of which rely on original optical flows.
			}
			\label{tb.depth_abl}  
			\begin{tabular}{l|cc|cc|cc}
				\toprule
				\multirow{2}{*}{Configuration} & \multicolumn{2}{c|}{KITTI Raw} & \multicolumn{2}{c|}{DDAD} & \multicolumn{2}{c}{nuScenes} \\
				\cline{2-7} 
				& Abs Rel & Sq Rel & Abs Rel & Sq Rel & Abs Rel & Sq Rel  \\
				\hline
				Lite-Mono  & 0.101 & 0.729 & 0.162 & 4.451 & 0.419 & 15.578  \\
				D-HRNet  & 0.102 & 0.760 & 0.198 & 8.124 & 0.396 & 15.087  \\
				\hline
				Baseline (Lite-Mono) & 0.104 & 0.759 & 0.150 & 3.255 &   0.160  &   7.250  \\
				Baseline (D-HRNet) & 0.104 & 0.810 & 0.155 & 3.408 &   0.160  &   7.188  \\
				\hline
				w/ \cite{zou2018df} (Lite-Mono) & 0.102 & 0.736 & 0.146 & 3.288 & 0.158 & 2.561  \\
				w/ \cite{zhang2024dcpi} (Lite-Mono) & 0.099 & 0.694 & 0.143 & 3.214 & 0.152 & 2.130  \\
				\hline
				w/ Ours (Lite-Mono) &\textbf{0.097} &\textbf{0.652} &\textbf{0.134} &2.685 &0.139 &1.472 \\
				w/ Ours (D-HRNet) &0.100 &0.729 &0.138 &\textbf{2.588} &\textbf{0.138} &\textbf{1.404} \\
				\bottomrule
			\end{tabular}
		\end{center}
	\end{table}

	The second ablation study investigates the effectiveness of the proposed constraints for refining DepthNet outputs, which are derived from decomposed flows following the alignment processes. While prior works have already explored the use of optical flow to constrain depth estimation with all motion types, we not only validate the effectiveness of our method but also compare it with previous approaches. The compared approaches include the cross-task consistency loss introduced in the study \cite{zou2018df}, which enforces consistency between rigid and optical flows, and the contextual-geometric depth consistency loss proposed in the study \cite{zhang2024dcpi}, where geometric depth is obtained via triangulation. As shown in Table~\ref{tb.depth_abl}, our proposed method significantly outperforms both prior arts across three datasets, supporting our claim that decomposing flows provides more effective geometric constraints. In addition, we replace the default DepthNet with the architecture proposed in \cite{he2022ra} to further evaluate the compatibility of DiMoDE. Despite the fundamental differences in architectural design, our method consistently delivers substantial performance gains, demonstrating its robustness and architectural flexibility.
	
	\begin{figure}[t!]
		\centering
		\includegraphics[width=0.48\textwidth]{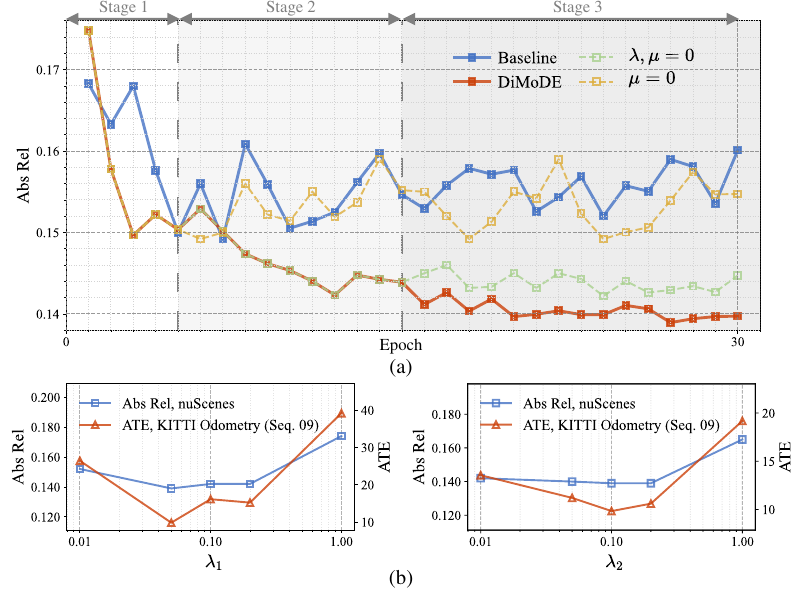}
		\caption{Quantitative results demonstrating DiMoDE’s convergence behavior and robustness to hyperparameter variations: (a) comparison of validation curves for the baseline model, DiMoDE, and its ablation variants; (b) analysis of the impact of varying hyperparameters $\lambda_1$ and $\lambda_2$ on the performance of both tasks.}
		\label{fig:convergence}
	\end{figure}
	
	Moreover, we analyze the evolution of the validation metric (Abs Rel) on the nuScenes dataset to analyze the impact of our method. 
	As shown in Fig.~\ref{fig:convergence}(a), the baseline model exhibits noticeable oscillations in its validation curve, indicating that photometric loss alone does not provide sufficiently effective supervision for depth estimation in relatively complex scenes. 
	In contrast, our method achieves significantly smoother convergence, emphasizing the effectiveness of the incorporated geometric constraints. 
	Moreover, we validate the effectiveness of our training strategy (see Sect.~\ref{Sect.implement}) through two ablation studies: one by removing $\mathcal{L}_\text{tan}$ and $\mathcal{L}_\text{rad}$ (\ie, setting $\lambda_2=0$) during epochs 16-30, and another by additionally removing $\mathcal{L}_\text{axi}$ and $\mathcal{L}_\text{pla}$ (\ie, setting $\lambda_1,\lambda_2=0$) throughout the entire training process. The results confirm that each component contributes to suppressing oscillations and improving overall performance.
	
	Finally, we conduct a hyperparameter selection experiment to determine the optimal weights in \eqref{eq.overall_loss}. As shown in Fig.~\ref{fig:convergence}(b), while the default setting ($\lambda_1=0.05$, $\lambda_2=0.1$) achieves the best performance across both tasks, adjusting the weights within a moderate range (from $0.05$ to $0.2$) yields comparable results. These results suggest that the performance is relatively robust to these hyperparameters. Nonetheless, performance degrades significantly when the weights are reduced to 0.01 or increased to 1. This degradation is attributed to insufficient gradient strength at lower values, which impedes effective optimization, and to the dominance of the geometric loss at higher values, which disrupts the balance with photometric supervision and leads to unstable training.

	\section{Discussion}
	\label{Sect.discussion}
	
	This section discusses two primary limitations of DiMoDE and analyzes their underlying causes.
	First, despite notable improvements over existing methods, DiMoDE still struggles to produce reliable predictions under extremely low illumination, as illustrated by the qualitative results on Indoor Seq. 01 in Fig.~\ref{fig.trajcreated}. This limitation primarily arises from the severe degradation of both photometric cues used in the reconstruction loss and the diminished contextual features for reliable dense correspondences generation.
	Second, although DiMoDE with enhanced constraints demonstrates more robust learning under challenging conditions, it fails to generalize effectively to other unseen real-world sequences. This limitation can be attributed to the significantly greater environmental and motion complexity present in our collected dataset compared to conventional driving scenarios. Consequently, the current network architecture and training strategies fall short in capturing generalizable patterns in such highly complex and diverse environments.

	\section{Conclusion and Future Works}
	\label{Sect.conclusion}
	This article presented DiMoDE, an unsupervised joint depth and ego-motion learning framework that discriminately treats motion components. 
	This study first revisited the rigid flows resulting from various motion types, uncovering their distinct characteristics as well as the long-overlooked limitations caused by indiscriminately mixing these flows when generating supervisory signals. To overcome these limitations, DiMoDE introduced explicit geometric constraints from motion components through the alignments of the optical axes and imaging planes between the source and target cameras. These constraints enable targeted optimization for each ego-motion component and enhance depth supervision. Extensive experiments conducted on six public datasets and a newly collected real-world dataset demonstrated that DiMoDE consistently achieves robust learning across diverse real-world environments, delivering impressive performance in both visual odometry and depth estimation tasks. 
	
	Future work will focus on addressing existing limitations, including enabling robust training and reliable predictions under extreme illumination, and enhancing generalizability in complex real-world environments through more interpretable motion representations. In addition, we plan to exploit unsupervised learning across diverse scenes for scalable model training, paving the way toward a new paradigm for developing depth foundation models.
	
	\normalem
	\bibliographystyle{IEEEtran}
	\bibliography{main}

\end{document}